\pdfoutput=1
\documentclass[11pt]{article}

\usepackage[preprint]{acl}

\usepackage{times}
\usepackage{latexsym}

\usepackage[T1]{fontenc}

\usepackage[utf8]{inputenc}

\usepackage{microtype}

\usepackage{inconsolata}

\usepackage{graphicx}
\usepackage{amsmath}
\usepackage{amssymb}
\usepackage{amsthm}
\newtheorem{proposition}{Proposition}
\usepackage{booktabs}
\usepackage{placeins}
\usepackage{colortbl}
\usepackage[most]{tcolorbox}
\usepackage{enumitem}
\usepackage{fontawesome5}
\usepackage{twemojis}
\definecolor{deltagreen}{RGB}{0,100,0}
\definecolor{deltared}{RGB}{139,0,0}
\definecolor{rowblue}{RGB}{222,235,247}
\definecolor{promptframe}{RGB}{52,73,94}
\newtcolorbox{promptbox}[1]{enhanced, colback=gray!4, colframe=promptframe,
  colbacktitle=promptframe, coltitle=white, fonttitle=\bfseries\small,
  title=#1, boxrule=0.6pt, arc=2pt, left=6pt, right=6pt, top=5pt,
  bottom=5pt, fontupper=\small}
\newcommand{\promptvar}[1]{\textcolor{blue!60!black}{\texttt{\{#1\}}}}
\newcommand{\deltapos}[1]{\textcolor{deltagreen}{\bfseries\boldmath #1}}
\newcommand{\deltaneg}[1]{\textcolor{deltared}{\bfseries\boldmath #1}}
\newcommand{\dpos}[1]{\ensuremath{{}_{\textcolor{deltagreen}{\scriptscriptstyle +#1}}}}
\newcommand{\dneg}[1]{\ensuremath{{}_{\textcolor{deltared}{\scriptscriptstyle -#1}}}}

\newcommand{\method}{\textsc{Re}\textsuperscript{3}\textsc{Rank}}

\title{\method: Listwise Reranking with  Reasoning-Conditioned \\ Retrieval Heads}

\author{Ruiyi Yan\thanks{This work was done during internship at LIGHTSPEED.} \\
  Kyoto University \\
  \texttt{ruiyi@nlp.ist.i.kyoto-u.ac.jp} \\\And
  Zhuoyuan Mao \\
  LIGHTSPEED \\
  \texttt{zhuoyuanmao@global.tencent.com} \\\AND
  Yugo Murawaki\thanks{Corresponding authors.} \\
  Kyoto University \\
  \texttt{murawaki@i.kyoto-u.ac.jp} \\\And
  Yiwen Guo\footnotemark[2] \\
  Independent Researcher \\
  \texttt{guoyiwen89@gmail.com} \\}

\begin{document}
\maketitle
{\renewcommand{\thefootnote}{}%
\footnotetext{The code is available at
\faGithub~\href{https://github.com/ryehr/RE3RANK}{github.com/ryehr/RE3RANK},\newline
and models at
\raisebox{-.15ex}{\twemoji[height=1.3\fontcharht\font`X]{1f917}}~\href{https://huggingface.co/collections/Ryehr/re3rank}{\mbox{huggingface.co/collections/Ryehr/re3rank}}.}}
\begin{abstract}
Attention-based rerankers score each candidate by the attention mass
that retrieval heads route to it from the question, but the question, serving as a fixed query, carries limited information for locating
relevant evidence. We propose \method{}, a listwise \textbf{re}ranker built on
\textbf{re}asoning-conditioned \textbf{re}trieval heads (RRHeads). A generator first
writes a reasoning trace over the question and the candidates, and a
scorer then uses the trace tokens as the attention query of a few
selected RRHeads, reading relevance as continuous scores.
We present a straight-through estimator to propagate the
ranking loss through the discrete trace, so that the generator and the
scorer can be trained jointly for ranking. Averaged over 9 benchmarks, \method{}
outperforms 11 baseline rerankers: our 4B variant exceeds the
strongest baseline by 3.11 NDCG@10 points on average, and even our
0.6B variant surpasses rerankers with up to 32B parameters.
\end{abstract}

\section{Introduction}
\label{sec:introduction}

Large language models (LLMs) are increasingly integrated into
retrieval-augmented generation (RAG) systems, which supply knowledge
beyond the model's parameters
\citep{jin2025searchr}.
Such systems
generally need to retrieve a candidate pool with a
\textit{retriever} and then reorder and filter it with a more expressive \textit{reranker}
\citep{doi:10.36227/techrxiv.176300630.01740917/v1}.

Advanced LLM-based rerankers commonly reason and verbalize relevance as a sequence of candidate identifiers~\citep{zhang-etal-2025-rearank,liu-etal-2026-reasonrank}, but
verbalization can distort the model's internal judgments~\citep{NEURIPS2023_81b83900, ICLR2025_b5b1890a}, and long generated outputs
slow down inference. A complementary line of work probes relevance
directly from internal attention. Retrieval heads locate relevant
information in long contexts \citep{wu2025retrieval}, and query-focused
retrieval heads (QRHeads) score each candidate by the attention mass it
receives from the question \citep{zhang-etal-2025-query-focused, li2026queryfocusedmemoryawarererankerlong}.

\addtocounter{footnote}{1}
\begin{figure}[t]
\centering
\includegraphics[width=\columnwidth]{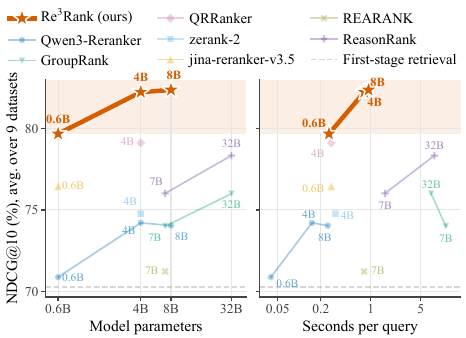}
\caption{NDCG@10 (\%) of rerankers averaged over 9 datasets, against
model size (left) and reranking time per query at a batch size of 50
(right).\protect\footnotemark[\value{footnote}]}
\label{fig:teaser-results}
\end{figure}
\footnotetext[\value{footnote}]{The size of \method{} refers to its
backbone. Its generator and scorer both have this size but run one
after the other, so the number of parameters activated at any time equals
the indicated size, which makes the comparison with the baselines fair
in terms of loaded parameters.}

\begin{figure*}[t]
\centering
\includegraphics[width=\textwidth]{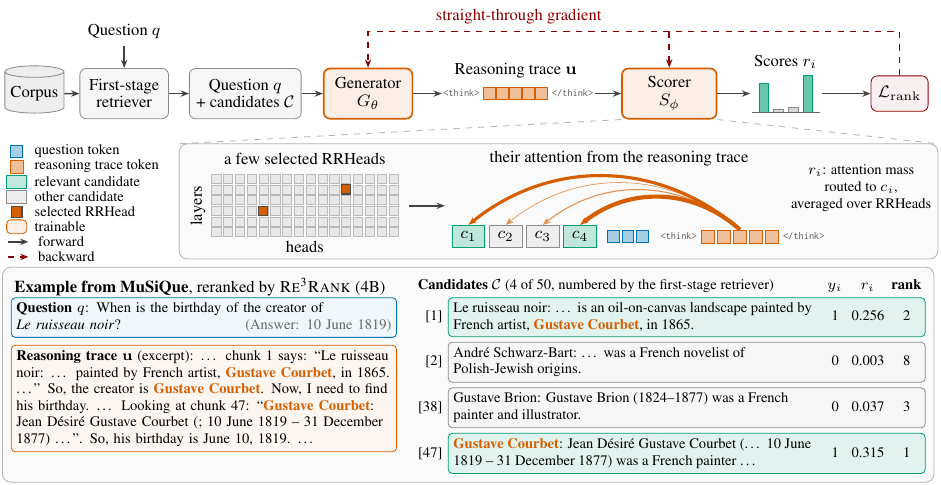}
\caption{Overview of \method{}. A first-stage retriever returns
candidates for the question from the corpus. The generator writes a
reasoning trace from the question and the candidates, and the scorer reads the same
input followed by the reasoning trace. Of all attention heads in the
scorer, only a few selected RRHeads are used: each candidate is scored
by the attention mass that these heads route from the reasoning trace
positions to it. The ranking loss
$\mathcal{L}_{\mathrm{rank}}$ updates both components and reaches the
generator through a straight-through estimator.}
\label{fig:teaser}
\end{figure*}

However, existing QRHead-based rerankers fix the attention query to
the tokens of the original question, which often lack the information
needed to locate relevant evidence.
Reasoning
rerankers do make intermediate reasoning steps explicit, but they still
convert their reasoning into a verbalized permutation.
Written after reading the candidates, such reasoning is more
expressive than the question, as it spells out information that the
question omits, such as bridge entities and intermediate facts. This
raises the question of whether it can replace the question as the
attention query.

Inspired by this, we propose \method{}, a listwise \textbf{re}ranker built on
{\textbf{re}asoning-conditioned \textbf{re}trieval heads} (RRHeads), as illustrated
in Figure~\ref{fig:teaser}. Given a question and its candidate pool, a
reasoning generator first writes a reasoning trace. A separate scorer then uses the
trace tokens as the attention queries
of selected RRHeads and scores each candidate by the attention mass it
receives. 

Our contributions are as follows.
First, we identify the attention query as a
dimension of attention-based rerankers and introduce RRHeads with a generated reasoning trace as the attention
query while scoring candidates continuously in one forward pass.
Second, we introduce a \textbf{straight-through joint training} method to train the reasoning generator and the scorer
jointly for ranking with a mixed ranking objective.
Third, we build three \method{} rerankers on Qwen3 backbones at 0.6B,
4B, and 8B \citep{yang2025qwen3technicalreport}, which
outperform 11 baseline rerankers on 9 benchmarks 
(Figure~\ref{fig:teaser-results}). The 4B variant exceeds QRRanker (4B) by
3.31 R@3 and 3.11 NDCG@10 on average, and even the 0.6B variant
outperforms every baseline, including 32B rerankers, in Recall@3, Recall@5, and
NDCG@10. 
Besides, \method{} can serve
as the retrieval component of a memory system, and
the methods of \method{} are compatible with
Gemma~4 backbones \citep{gemmateam2026gemma4}. 

\section{Preliminaries}
\label{sec:preliminaries}

\subsection{Post-Retrieval Reranking}
\label{sec:reranking}

Retrieval-augmented generation (RAG) supplements a language model's
parametric knowledge with non-parametric evidence retrieved from an
external corpus, providing explicit context for knowledge-intensive
reasoning and grounded generation while helping mitigate hallucinations
\citep{NEURIPS2020_6b493230,pmlr-v119-guu20a,izacard-grave-2021-leveraging,li2024enhancingllmfactualaccuracy}.
In practical RAG pipelines, an efficient first-stage retriever constructs
a candidate pool, after which a more expressive second-stage reranker can
refine its ordering before generation
\citep{karpukhin-etal-2020-dense,nogueira2020passagererankingbert,glass-etal-2022-re2g,chuang-etal-2023-expand,doi:10.36227/techrxiv.176300630.01740917/v1}.
Let $q$ denote a question and
$\mathcal{C}=\{c_1,\ldots,c_C\}$ the candidate set returned by a
first-stage retriever. For training and evaluation, each candidate
carries a gold binary relevance label $y_i\in\{0,1\}$, collected in
$\mathbf{y}=(y_1,\ldots,y_C)$; the labels come mostly from dataset
annotations (Appendix~\ref{app:labels}) and are not available to the
reranker at inference. We write
$\mathcal{P}=\{i\mid y_i=1\}$ and
$\mathcal{N}=\{i\mid y_i=0\}$ for the indices of the in-pool relevant
and irrelevant candidates, respectively.

A reranker maps the question and the candidate set to one
real-valued score per candidate and orders the candidates by
decreasing score:
\begin{equation}
\label{eq:reranking}
 \mathbf{r}
 = f_{\psi}(q,\mathcal{C})\in\mathbb{R}^{C}.
\end{equation}
Here, $\psi$ denotes reranker parameters.
Let $\mathcal{D}$ denote the distribution over labeled reranking
instances $(q,\mathcal{C},\mathbf{y})$. For a target cutoff $k$, a reranking objective is to maximize the conditional Recall@$k$ (R@$k$) over the candidate pool:
\begin{equation}
\label{eq:reranking-objective}
\max_{\psi}\;
\mathbb{E}_{(q,\mathcal{C},\mathbf{y})\sim\mathcal{D}}
\left[
\frac{1}{|\mathcal{P}|}
\sum_{i\in\mathcal{P}}
\mathbb{I}\!\left\{i\in\operatorname{Top}_{k}(\mathbf{r})\right\}
\right].
\end{equation}
Here, $\operatorname{Top}_{k}(\mathbf{r})$ denotes the indices of the
$k$ highest scores, and normalization by $|\mathcal{P}|$ restricts
recall to relevant candidates present in the candidate pool.

\subsection{Reranking Paradigms}
\label{sec:reranking-paradigms}

Rerankers are commonly categorized by how many candidates they compare
at once. 
(1) \emph{Pointwise} methods score each query--candidate pair
independently \citep{nogueira2020passagererankingbert,10.1145/3805712.3809850,chen2026polarizationgenerativeconstraintchainofthought}. They scale
linearly with the pool size but cannot use other candidates as context
for an individual decision. 
(2) \emph{Pairwise} methods instead predict a
preference between two candidates and aggregate such local comparisons
into a global order \citep{qin-etal-2024-large,10.1145/3746252.3760859}. 
(3) \emph{Listwise} methods expose the model to multiple candidates
and output either a permutation or a score vector
\citep{sun-etal-2023-chatgpt,gangi-reddy-etal-2024-first, 10.1145/3696410.3714658,yan2026memdefraglatentmemorydefragmentation}. This enables
cross-candidate reasoning, at the cost of longer prefilled contexts.
Recent listwise rerankers further generate a reasoning chain before
the permutation \citep{zhang-etal-2025-rearank,liu-etal-2026-reasonrank}. 
\citet{zhang-etal-2025-query-focused} and
\citet{li2026queryfocusedmemoryawarererankerlong}
show that listwise rerankers based on query-focused retrieval heads
(QRHeads) can achieve state-of-the-art performance by extracting
relevance directly from internal attention patterns. Such internal
probing, rather than verbalized relevance judgments or generated
permutations, offers a promising direction for LLM-based reranking.

\subsection{Query-Focused Retrieval Heads (QRHeads)}
\label{sec:qrhead}

Consider a decoder-only Transformer applied to a sequence
in which the candidate contexts precede the question. At layer $l$ and
head $h$, let $Q^{(l,h)}_t$ and $K^{(l,h)}_j$ be the query and key
vectors at positions $t$ and $j$, and let $d_h$ be the head dimension.
The causally masked attention weight is
$A^{(l,h)}_{t,j}=\operatorname{softmax}_{j\le t}\bigl(Q^{(l,h)}_t\cdot
K^{(l,h)}_j/\sqrt{d_h}\bigr)$, normalized over the positions $j\le t$,
with $A^{(l,h)}_{t,j}=0$ for future positions $j>t$. Let
$\mathcal{I}_i$ denote the contiguous span of positions occupied by the
tokens of candidate $c_i$ in the sequence, and let
$\mathcal{Q}$ denote a set of attention-query positions, such as the
positions $\mathcal{Q}_q$ occupied by the tokens of $q$.
Since the candidates precede $\mathcal{Q}$, every $j\in\mathcal{I}_i$
satisfies $j<t$ for all $t\in\mathcal{Q}$. The attention mass routed from $\mathcal{Q}$
to candidate $c_i$ is
\begin{equation}
\label{eq:head-mass}
 m_i^{(l,h)}(\mathcal{Q})
 =
 \frac{1}{|\mathcal{Q}|}
 \sum_{t\in\mathcal{Q}}
 \sum_{j\in\mathcal{I}_i}
 A^{(l,h)}_{t,j}.
\end{equation}

\citet{wu2025retrieval} identify retrieval heads as attention heads
associated with locating relevant information in long contexts.
Building on this idea,
\citet{zhang-etal-2025-query-focused} and
\citet{li2026queryfocusedmemoryawarererankerlong}
select query-focused retrieval heads (QRHeads) whose query-to-context attention
predicts relevance on a small annotated set. Let $\mathcal{H}_{\mathrm{QR}}$ denote the selected
heads. A QRHead reranker sets $\mathcal{Q}=\mathcal{Q}_q$ and
instantiates Eq.~\ref{eq:reranking} by averaging
Eq.~\ref{eq:head-mass} over these heads, for
$i=1,\ldots,C$:
\begin{equation}
\label{eq:qr-score}
r_i =
\frac{1}{|\mathcal{H}_{\mathrm{QR}}|}
\sum_{(l,h)\in\mathcal{H}_{\mathrm{QR}}}
m_i^{(l,h)}(\mathcal{Q}_q).
\end{equation}
The score vector $\mathbf{r}=(r_1,\ldots,r_C)$ is thus read directly
from attention, without decoding document identifiers or a ranking
permutation.

\section{\texorpdfstring{\method}{Re3Rank}}
\label{sec:method}

Existing QRHead-based rerankers optimize head selection or the scoring
model while fixing the attention query to $\mathcal{Q}_q$. They
therefore treat what serves as the attention query as immutable,
overlooking the optimization space of constructing query content that
better routes attention toward relevant candidates. This work addresses this
neglected dimension by replacing the fixed question positions with a
model-generated reasoning trace, making query construction itself
trainable. Retrieval heads used in this way are termed
\emph{reasoning-conditioned retrieval heads} (RRHeads).

\subsection{Overview}
\label{sec:method-overview}

\method{} comprises a reasoning generator $G_{\theta}$ and a separate
attention-based scorer $S_{\phi}$. They are initialized from the same
backbone. Given a question and its candidate pool (retrieved from the first-stage retriever),
$G_{\theta}$ first produces a reasoning trace. $S_{\phi}$ then uses the
trace-token positions $\mathcal{Q}_u$ in place of $\mathcal{Q}_q$ in
Eq.~\ref{eq:head-mass} and
computes continuous candidate scores. To optimize the generator despite
the discrete trace, we connect the two models with a straight-through
relaxation, allowing the selected ranking objective to jointly
train both components.

\subsection{Reasoning Trace Generation and Scoring}
\label{sec:trace-scoring}

We denote the generator prompt as the token sequence
$\mathbf{x}=\operatorname{TokenPrompt}(\mathcal{C}\mathbin{\|}q)
=(x_1,\ldots,x_n)$, where $\mathbin{\|}$ denotes ordered
concatenation. $\operatorname{TokenPrompt}$ preserves the candidate order, and
applies the backbone model's native instruction format.
The generator $G_{\theta}$ continues the prompt with a reasoning trace
$\mathbf{u}=(u_1,\ldots,u_T)$, whose tokens are sampled autoregressively:
\begin{equation}
\label{eq:trace-policy}
u_t\sim p_{\theta}(\,\cdot\mid\mathbf{x},\mathbf{u}_{<t}).
\end{equation}
Here, $\mathbf{u}_{<t}=(u_1,\ldots,u_{t-1})$.

The trace occupies query positions
$\mathcal{Q}_u=\{n+1,\ldots,n+T\}$. Using Eq.~\ref{eq:head-mass} to
compute the attention mass from $\mathcal{Q}_u$ with a selected RRHead
set $\mathcal{H}_{\mathrm{RR}}$, Eq.~\ref{eq:qr-score} becomes
\begin{equation}
\label{eq:rr-score}
r_i
=
\frac{1}{|\mathcal{H}_{\mathrm{RR}}|}
\sum_{(l,h)\in\mathcal{H}_{\mathrm{RR}}}
m_i^{(l,h)}(\mathcal{Q}_u).
\end{equation}
$\mathcal{H}_{\mathrm{RR}}$ is identified on a small labeled seed set by ranking
heads according to the attention mass they assign to relevant
candidates; Section~\ref{sec:rrhead-selection} gives the selection
criterion and the resulting heads.

The procedure above also defines the inference procedure:
$G_{\theta}$ generates a trace $\mathbf{u}$, after which
$S_{\phi}$ uses $\mathcal{Q}_u$ as its attention-query positions,
computes Eq.~\ref{eq:rr-score}, and sorts candidates by $r_i$.

\subsection{Ranking Objectives}
\label{sec:ranking-objective}

Raw attention masses vary in scale across examples. The scores of each
candidate pool are first normalized to $[0,\alpha]$:
\begin{equation}
\label{eq:maxmin}
\widetilde r_i
=
\alpha\,
\frac{r_i-\min_j r_j}
{\max_j r_j-\min_j r_j+\epsilon},
\end{equation}
where $\alpha>0$ controls the score scale and $\epsilon>0$ ensures
numerical stability. In the default setting, they are instantiated with $8$ and
$10^{-6}$, respectively.

\subsubsection{Contrastive Cross-Entropy Loss}
\label{sec:contrastive-objective}

The primary objective is a grouped contrastive cross-entropy:
\begin{equation}
\label{eq:ce-loss}
\begin{aligned}
\mathcal{L}_{\mathrm{CE}}
&=
-\frac{1}{|\mathcal{P}|}
\sum_{p\in\mathcal{P}}
\log\frac{\exp(\widetilde r_p)}{Z_p},\\
Z_p
&=
\exp(\widetilde r_p)
+\sum_{j\in\mathcal{N}}\exp(\widetilde r_j).
\end{aligned}
\end{equation}
Each relevant candidate is contrasted against every irrelevant
candidate in the same pool, and examples with $\mathcal{P}=\varnothing$
are excluded. However, the loss keeps enlarging margins that are
already correct, spending gradient on relevant candidates that outrank every
irrelevant one rather than on those still outranked
\citep{Sun_2020_CVPR, patel2022recall}.

\subsubsection{\texorpdfstring{Recall@$k$ Surrogate}{Recall@k Surrogate} Loss}
\label{sec:recall-surrogate}

Recall@$k$, in contrast, depends only on whether relevant candidates
cross the top-$k$ boundary. Inspired by \citet{patel2022recall}, we
target this boundary directly by relaxing the discrete rank and the
top-$k$ indicator with the tempered sigmoid
\begin{equation}
\label{eq:tempered-sigmoid}
\sigma_{\tau}(a)
=
\frac{1}{1+\exp(-a/\tau)}.
\end{equation}
For each relevant candidate $p\in\mathcal{P}$, its soft rank and the
soft number of relevant candidates within the top $k$ are
\begin{align}
\rho_p
&=
1+\sum_{\substack{z=1\\z\ne p}}^{C}
\sigma_{\tau_2}(\widetilde r_z-\widetilde r_p),
\label{eq:soft-rank}\\
h_k
&=
\min\!\left\{
\sum_{p\in\mathcal{P}}\sigma_{\tau_1}(k-\rho_p),
\min(k,|\mathcal{P}|)
\right\}.
\label{eq:soft-hits}
\end{align}
Here, $\tau_2$ controls the rank relaxation and $\tau_1$ controls the
top-$k$ boundary. Clipping $h_k$ and normalizing by
$\min(k,|\mathcal{P}|)$ account for the maximum attainable number of
hits. For a set of cutoffs $\mathcal{K}$, the Recall@$k$ surrogate is
\begin{equation}
\label{eq:recall-surrogate}
\mathcal{L}_{\mathrm{rsk}}
=
\frac{1}{|\mathcal{K}|}
\sum_{k\in\mathcal{K}}
\left(
1-\frac{h_k}{\min(k,|\mathcal{P}|)}
\right).
\end{equation}
The surrogate can supplement the contrastive:
\begin{equation}
\label{eq:mixed-ranking-loss}
\mathcal{L}_{\mathrm{rank}}
=
\mathcal{L}_{\mathrm{CE}}
+\lambda\mathcal{L}_{\mathrm{rsk}},
\qquad \lambda\ge0.
\end{equation}
\subsection{Straight-Through Joint Training}
\label{sec:straight-through}

The generator and scorer define the joint stochastic objective
\begin{equation}
\label{eq:joint-objective}
\begin{aligned}
\overline{\mathcal L}_{\theta,\phi}(\mathbf{x},\mathbf y)
&=
\mathbb{E}_{\mathbf{u}\sim p_{\theta}(\cdot\mid\mathbf{x})}
\bigl[
\mathcal{L}_{\mathrm{rank}}
(S_{\phi}(\mathbf{x},\mathbf{u}),\mathbf y)
\bigr],\\
\mathcal{J}(\theta,\phi)
&=
\mathbb{E}_{(q,\mathcal{C},\mathbf y)\sim\mathcal D}
\bigl[
\overline{\mathcal L}_{\theta,\phi}(\mathbf{x},\mathbf y)
\bigr].
\end{aligned}
\end{equation}
Here, $S_{\phi}(\mathbf{x},\mathbf{u})$ denotes the score vector in
Eq.~\ref{eq:rr-score}.
The sampled tokens in Eq.~\ref{eq:trace-policy} block direct pathwise
differentiation with respect to $\theta$. We therefore use a
straight-through estimator \citep{bengio2013estimatingpropagatinggradientsstochastic}. After one hard
trace is sampled without gradients, $G_{\theta}$ is run once under
teacher forcing on $\mathbf{x}\mathbin{\|}\mathbf{u}$. Let $V$ and $d$
denote the shared vocabulary size and scorer hidden dimension,
respectively, and let $E\in\mathbb{R}^{V\times d}$ be the scorer's
frozen token-embedding matrix. The logits at position $n+t-1$, which
predict $u_t$, and the corresponding relaxed token distribution are
\begin{equation}
\label{eq:soft-token}
\begin{aligned}
\boldsymbol{\ell}_t
&=G_{\theta}(\mathbf{x}\mathbin{\|}\mathbf{u})_{n+t-1}
\in\mathbb{R}^{V},\\
\mathbf p_t
&=\operatorname{softmax}(\boldsymbol{\ell}_t/\tau_{\mathrm{ST}}).
\end{aligned}
\end{equation}
Here, $\tau_{\mathrm{ST}}>0$ is the straight-through relaxation
temperature.
The scorer embedding at trace position $t$ is constructed as
\begin{equation}
\label{eq:straight-through}
\widetilde{\mathbf e}_t
=
E[u_t]
+E^{\top}\mathbf p_t
-\operatorname{sg}\!\left(E^{\top}\mathbf p_t\right),
\end{equation}
where $\operatorname{sg}(\cdot)$ denotes stop-gradient. Numerically,
the two relaxed terms cancel in the forward pass, so
$\widetilde{\mathbf e}_t=E[u_t]$. In the backward pass, however,
\begin{equation}
\label{eq:st-backward}
\begin{aligned}
\frac{\partial\widetilde{\mathbf e}_t}
{\partial\boldsymbol{\ell}_t}
&=E^{\top}\mathbf J_t,\\
\mathbf J_t
&=\frac{1}{\tau_{\mathrm{ST}}}
\left[
\operatorname{Diag}(\mathbf p_t)
-\mathbf p_t\mathbf p_t^{\top}
\right].
\end{aligned}
\end{equation}
The scorer's embedding gradient therefore becomes a token-level signal
for the generator through the softmax Jacobian. This estimator is
biased, as it substitutes a relaxed backward path for the
derivative of discrete sampling, but it provides a low-variance
differentiable route into $G_{\theta}$. A single backward pass updates both
$G_{\theta}$ and $S_{\phi}$.\footnote{Appendix~\ref{app:st-discussion} details a discussion of the implications of
this update for trace generation.}

\section{Experiments}
\label{sec:experiments}

\subsection{RRHead Selection}
\label{sec:rrhead-selection}

RRHeads were selected independently for each Qwen3 backbone scale
\citep{yang2025qwen3technicalreport} before training. The base model
first greedily generated a reasoning trace for each instance in a seed
set $\mathcal{D}_{\mathrm{seed}}$ of 800 labeled instances. Every attention head of the
backbone ($28$ layers $\times$ $16$ heads for 0.6B and $36\times32$
for 4B and 8B) was then scored by the average attention mass that the
trace positions route to relevant candidates:
\begin{equation}
\label{eq:head-selection}
\gamma^{(l,h)}
=
\frac{1}{|\mathcal{D}_{\mathrm{seed}}|}
\sum_{(q,\mathcal{C},\mathbf{y})\in\mathcal{D}_{\mathrm{seed}}}
\sum_{i\in\mathcal{P}}
m_i^{(l,h)}(\mathcal{Q}_u),
\end{equation}
where $\mathcal{P}$, $\mathcal{Q}_u$, and $m_i^{(l,h)}$ were computed
for each instance as in Section~\ref{sec:trace-scoring}. The two
highest-scoring heads formed $\mathcal{H}_{\mathrm{RR}}$. The selected sets were $\{(20,4),(16,8)\}$ for
Qwen3-0.6B, $\{(23,10),(21,8)\}$ for Qwen3-4B, and
$\{(15,11),(23,10)\}$ for Qwen3-8B. {Appendix~\ref{app:rrhead-selection}
details this procedure for Qwen3-4B.}

\subsection{Implementations of Training}
\label{sec:training-data}

The training mixture contained 38,000 English questions drawn from
NarrativeQA \citep{10.1162/tacl_a_00023}, MuSiQue
\citep{10.1162/tacl_a_00475}, ConvoMem
\citep{pakhomov2025convomembenchmark150conversations}, QAConv
\citep{wu-etal-2022-qaconv}, and FriendsQA
\citep{yang-choi-2019-friendsqa}. 
Each instance contained a question, for which up to 50 candidates were
retrieved by Qwen3-Embedding-8B~\citep{zhang2025qwen3embeddingadvancingtext}
before reranking. All constructed instances contained at least one
relevant candidate. Table~\ref{tab:training-data} summarizes the
number of questions and in-pool relevant candidates per
question, and mean prompt length for each source.\footnote{Appendix~\ref{app:hyperparams} gives the training hyperparameters.}

\begin{table}[t]
\centering
\small
\begin{tabular*}{\columnwidth}{@{\extracolsep{\fill}}lrrr@{}}
\toprule
\textbf{Source} & \textbf{\#Questions} & \textbf{Avg. Gold} &
\textbf{Avg. Tokens} \\
\midrule
NarrativeQA & 12,000 & 2.47 & 10,213 \\
MuSiQue     & 12,000 & 1.83 &  7,378 \\
ConvoMem    &  7,000 & 2.06 & 10,390 \\
QAConv      &  5,000 & 1.05 &  9,857 \\
FriendsQA   &  2,000 & 1.01 & 11,940 \\
\midrule
Total       & 38,000 & 1.93 &  9,394 \\
\bottomrule
\end{tabular*}
\caption{Statistics of the training mixture by source dataset. Avg. Gold is
the mean number of in-pool relevant candidates per question; Avg.
Tokens is the mean prompt length before the generated trace.}
\label{tab:training-data}
\end{table}

\subsection{Implementations of Evaluation}
\label{sec:evaluation}

\subsubsection{Benchmarks}
\label{sec:benchmarks}

\begin{figure*}[t]
  \centering
  \includegraphics[width=\textwidth]{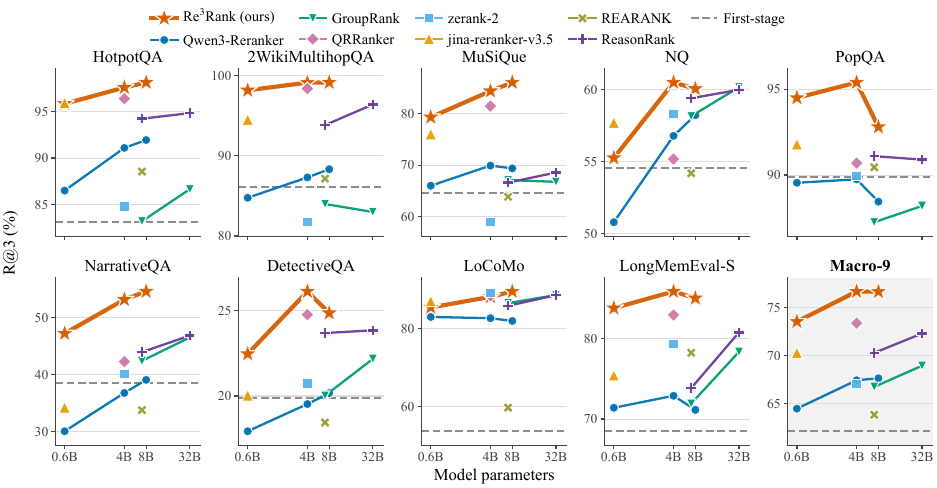}
  \caption{Conditional R@3 (\%) of rerankers across model sizes on the
  nine benchmarks and their macro average (Macro-9). Lines connect sizes of the same
  model family, and the dashed line marks the first-stage ranking of
  Qwen3-Embedding-8B.}
  \label{fig:r3-vs-size}
\end{figure*}

We evaluated on nine English test sets that span four retrieval regimes:
Wikipedia multi-hop QA with HotpotQA \citep{yang-etal-2018-hotpotqa},
2WikiMultihopQA \citep{ho-etal-2020-constructing}, and MuSiQue
\citep{10.1162/tacl_a_00475}; open-domain factoid QA with Natural
Questions (NQ) \citep{kwiatkowski-etal-2019-natural} and PopQA
\citep{mallen-etal-2023-trust}; long-narrative QA with the NarrativeQA
\citep{10.1162/tacl_a_00023} subset of HELMET \citep{yen2025helmetevaluatelongcontextlanguage} and
the English portion of DetectiveQA
\citep{xu2025detectiveqa}; and long-term
conversational memory with LoCoMo \citep{maharana-etal-2024-evaluating}
and LongMemEval-S \citep{wu2025longmemeval}. For every
question, the first-stage retriever of Section~\ref{sec:training-data}
returned the top 50 candidates from a dataset-specific retrieval scope,
and all rerankers scored the same candidate pools.\footnote{Appendix~\ref{app:corpora} describes the corpora and chunking.}

\subsubsection{Baselines}
\label{sec:baselines}

\method{} was compared with rerankers from the paradigms introduced in
Section~\ref{sec:reranking-paradigms}: the QRHead-based listwise
reranker QRRanker \citep{li2026queryfocusedmemoryawarererankerlong},
built on Qwen3-4B; the pointwise rerankers Qwen3-Reranker
\citep{zhang2025qwen3embeddingadvancingtext} at 0.6B, 4B, and 8B and
zerank-2 \citep{pipitone2025zeloeloinspiredtrainingmethod} at 4B; the
listwise reranker jina-reranker-v3.5
\citep{nasika2026jinarerankerv35efficientlistwisereranker} at 0.6B; the
reasoning listwise rerankers REARANK \citep{zhang-etal-2025-rearank} at
7B and ReasonRank \citep{liu-etal-2026-reasonrank} at 7B and 32B; and
the generative groupwise reranker GroupRank
\citep{long-etal-2026-grouprank} at 7B and 32B. All
methods reranked the same top-50 candidate pools.

\subsection{Main Results}
\label{sec:main-results}

\paragraph{Configuration.}
Unless stated otherwise, $\mathcal{L}_{\mathrm{rank}}$ in
Eq.~\ref{eq:mixed-ranking-loss} used $\lambda=2$ throughout the paper,
with cutoffs $\mathcal{K}=\{3,10\}$ and temperatures $\tau_1=\tau_2=1$
for the Recall@$k$ surrogate in
Eqs.~\ref{eq:soft-rank}--\ref{eq:recall-surrogate}. \method{} variants
in this section were trained with this objective. The three scales shared the training recipe of
Section~\ref{sec:training-data} and differed only in the backbone and
$\mathcal{H}_{\mathrm{RR}}$. At inference, the generator decoded
greedily with at most 1,024 trace tokens.

\paragraph{Overall comparison.}
Figure~\ref{fig:r3-vs-size} compares the R@3 of rerankers across model
sizes. Appendix~\ref{app:full-results} defines R@3 and normalized
discounted cumulative gain (NDCG@10;~\citealp{jarvelin2002cumulated}),
shows the corresponding NDCG@10 comparison in
Figure~\ref{fig:ndcg10-vs-size}, and reports all results in
Tables~\ref{tab:main-wiki}--\ref{tab:main-dialogue}. 
With the 4B backbone, \method{} reaches 76.70 R@3 and 82.23 NDCG@10 on
Macro-9, exceeding the strongest baseline, QRRanker, by 3.31 and 3.11
points and the first-stage ranking by 14.61 and 11.96 points. 

\paragraph{Parameter efficiency.}
At each scale, \method{} outperforms the size-matched Qwen3-Reranker by
9.05--9.28 R@3 and 8.02--8.79 NDCG@10 points on Macro-9. The 0.6B
variant already achieves higher Macro-9 R@3, R@5, and NDCG@10 than
every baseline, including GroupRank-32B and ReasonRank-32B; it exceeds ReasonRank-32B
by 1.24 R@3 and 1.33 NDCG@10 points. Scaling \method{} from 0.6B to 4B
adds 3.15 R@3 and 2.56 NDCG@10 points.

\paragraph{Comparison with reasoning rerankers.}
REARANK and ReasonRank also reason over the candidates before ranking,
but output the ranking as a generated permutation. The stronger ReasonRank-32B reaches 72.31 R@3 and 78.34 NDCG@10 on Macro-9,
4.39 and 3.89 points below \method{} (4B). This gap indicates effectiveness of
probing relevance from internal attention compared to verbalizing the model's relevance
judgments as a discrete sequence of candidate identifiers which can
distort them in the process.

\begin{figure}[t]
\centering
\includegraphics[width=\columnwidth]{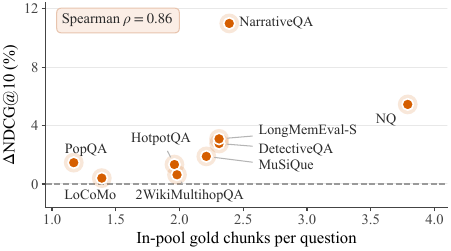}
\caption{NDCG@10 gain of \method{} (4B) over QRRanker on each
benchmark, measured as a difference and shown with the mean
number of in-pool relevant chunks per question from
Table~\ref{tab:benchmarks}.}
\label{fig:delta-gold}
\end{figure}

\paragraph{Comparison with QRRanker.}
QRRanker shares the attention-mass scoring of Eq.~\ref{eq:head-mass}
and the Qwen3-4B backbone with \method{} (4B) but uses the question as
the attention query. \method{} (4B) achieves higher NDCG@10 than
QRRanker on all nine datasets and higher R@3 on eight, with the largest
R@3 gains on NarrativeQA (+10.89), NQ (+5.32), and PopQA (+4.70).
Figure~\ref{fig:delta-gold} further shows that
the NDCG@10 gain grows with the number of in-pool relevant chunks per
question (Spearman $\rho=0.86$ across the nine datasets). The two
datasets with the fewest such chunks, PopQA and LoCoMo, gain less than
1.5 points, whereas NQ, which has the most, gains 5.45 points;
NarrativeQA is an outlier with an 11.00-point gain. This trend is
consistent with reasoning traces helping most when relevant evidence
spans several chunks.

\subsection{Ablation Study}
\label{sec:ablation}

\subsubsection{Ablation Study of Query Construction}
\label{sec:ablation-query}

Table~\ref{tab:ablation-query} compares query construction on Qwen3-4B.
QR denotes query-focused retrieval with QRHeads
(Section~\ref{sec:qrhead}), where a single model scores candidates with
the question tokens as attention queries. RR denotes
reasoning-conditioned retrieval with RRHeads
(Section~\ref{sec:method}), where the attention queries are the
reasoning trace generated by a separate model. Without training, RR already improves over QR by 2.57
R@3 and 1.40 NDCG@10 points.
After both were trained with $\mathcal{L}_{\mathrm{rank}}$, RR outperforms
QR on all four metrics, by 1.44 R@1, 2.65 R@3, 2.03 R@5, and 1.90
NDCG@10 points. Appendix~\ref{app:trace-content} analyzes which parts
of the trace contribute to this advantage.

\subsubsection{Ablation Study of Joint Training and Ranking Objectives}
\label{sec:ablation-joint}

\begin{table}[t]
  \centering
  \footnotesize
  \setlength{\tabcolsep}{2pt}
  \begin{tabular*}{\columnwidth}{@{\extracolsep{\fill}}lcccc@{}}
  \toprule
  \textbf{Method} & \textbf{R@1} & \textbf{R@3} & \textbf{R@5} & \textbf{N@10}\\
  \midrule
  QR (base)    & 41.14\hphantom{\dpos{0.00}} & 69.16\hphantom{\dpos{0.00}} & 77.06\hphantom{\dpos{0.00}} & 76.12\hphantom{\dpos{0.00}}\\
  \textbf{RR (base)} & \textbf{41.15}\dpos{0.01} & \textbf{71.73}\dpos{2.57} & \textbf{79.60}\dpos{2.54} & \textbf{77.52}\dpos{1.40}\\
  \midrule
  QR (trained) & 42.76\hphantom{\dpos{0.00}} & 74.05\hphantom{\dpos{0.00}} & 81.98\hphantom{\dpos{0.00}} & 80.33\hphantom{\dpos{0.00}}\\
  \textbf{RR (trained)} & \textbf{44.20}\dpos{1.44} & \textbf{76.70}\dpos{2.65} & \textbf{84.01}\dpos{2.03} & \textbf{82.23}\dpos{1.90}\\
  \bottomrule
  \end{tabular*}
  \caption{Comparison of query construction for Qwen3-4B, reported as
  Macro-9 scores (\%). QR uses a single model with the question tokens as
  attention queries and its top-2 QRHeads, $(23,10)$ and $(15,11)$. RR
  uses the generator's reasoning trace as attention queries with the
  RRHeads $(23,10)$ and $(21,8)$. Base rows kept all weights untrained,
  and trained rows were trained with $\mathcal{L}_{\mathrm{rank}}$ in
  Eq.~\ref{eq:mixed-ranking-loss}. Subscripts give the gain of RR over QR
  under the same training setting.}
  \label{tab:ablation-query}
  \end{table}

\begin{figure}[t]
\centering
\includegraphics[width=\columnwidth]{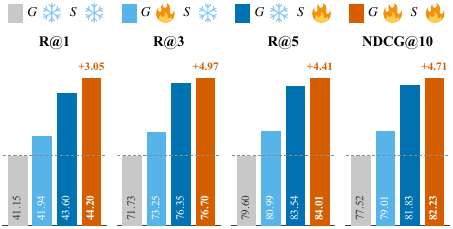}
\caption{Ablation of joint training for \method{} (4B) trained with
$\mathcal{L}_{\mathrm{rank}}$, reported as Macro-9 scores (\%). Four settings refer to whether the generator ($G$) and the scorer
($S$) were trained or kept at their base weights.}
\label{fig:ablation-joint}
\end{figure}

Figure~\ref{fig:ablation-joint} isolates the contributions of training the generator and
the scorer (4B models). Without any training, RRHeads computed on
base-model reasoning traces reach 71.73 R@3. Training only the generator,
with the scorer frozen, raises R@3 by 1.52 points, and training only the
scorer, with the generator frozen, raises it by 4.62 points, which means that training
the scorer is more effective than training the generator. Joint
training achieves the best result on all four metrics and improves over
scorer-only training by 0.35 R@3 and 0.40 NDCG@10 points, indicating
that the generator and scorer benefit from being optimized together.

\begin{table}[t]
\centering
\footnotesize
\setlength{\tabcolsep}{1pt}
\begin{tabular*}{\columnwidth}{@{\extracolsep{\fill}}lcccc@{}}
\toprule
\textbf{Objective} &
\textbf{R@1} & \textbf{R@3} & \textbf{R@5} & \textbf{N@10}\\
\midrule
$\mathcal{L}_{\mathrm{CE}}$ & 43.91\hphantom{\dpos{0.00}} & 76.57\hphantom{\dpos{0.00}} & 83.87\hphantom{\dpos{0.00}} & 82.05\hphantom{\dpos{0.00}}\\ \midrule
$\mathcal{L}_{\mathrm{CE}}$, $\alpha{=}12$ & 43.77\dneg{0.14} & 75.60\dneg{0.97} & 83.09\dneg{0.78} & 81.35\dneg{0.70}\\
$\mathcal{L}_{\mathrm{rsk}}$ & 43.79\dneg{0.12} & \textbf{76.81}\dpos{0.24} & 83.75\dneg{0.12} & 81.99\dneg{0.06}\\
$\mathcal{L}_{\mathrm{CE}}{+}\mathcal{L}_{\mathrm{rsk}}$ & \textbf{44.53}\dpos{0.62} & 76.49\dneg{0.08} & 83.72\dneg{0.15} & \textbf{82.28}\dpos{0.23}\\
$\mathcal{L}_{\mathrm{CE}}{+}2\mathcal{L}_{\mathrm{rsk}}$\,$^{\dagger}$ & 44.20\dpos{0.29} & 76.70\dpos{0.13} & \textbf{84.01}\dpos{0.14} & 82.23\dpos{0.18}\\
$\mathcal{L}_{\mathrm{CE}}{+}4\mathcal{L}_{\mathrm{rsk}}$ & 44.15\dpos{0.24} & 76.62\dpos{0.05} & 83.98\dpos{0.11} & 82.19\dpos{0.14}\\
\bottomrule
\end{tabular*}
\caption{Ablation of ranking objectives for \method{} (4B), reported as
Macro-9 scores (\%). $\mathcal{L}_{\mathrm{rsk}}$ used
$\mathcal{K}=\{3,10\}$ and $\tau_1=\tau_2=1$, and all rows used the score
scale $\alpha=8$ in Eq.~\ref{eq:maxmin} unless noted. Subscripts give
the difference from the first row, with gains in green and losses in
red. $^{\dagger}$Main configuration. Bold marks the best result in each
column.}
\label{tab:ablation-loss}
\end{table}

Table~\ref{tab:ablation-loss} compares ranking objectives for the 4B
model. The Recall@$k$ surrogate $\mathcal{L}_{\mathrm{rsk}}$ alone
yields the highest R@3 but slightly lowers the other metrics, whereas
adding it to $\mathcal{L}_{\mathrm{CE}}$ gives higher R@1 and NDCG@10
than either objective alone for every weight $\lambda$. With
$\lambda=2$, $\mathcal{L}_{\mathrm{rank}}$ improves over
$\mathcal{L}_{\mathrm{CE}}$ on all four metrics.
Appendix~\ref{app:ablation-loss} analyzes these results in detail,
including the effect of the score scale $\alpha$.

Appendix~\ref{app:rl} compares straight-through training with
reinforcement learning, which yields no notable gain over scorer-only
training.

\subsection{\texorpdfstring{\method{}}{Re3Rank} as a Memory Retrieval Component}
\label{sec:performance-as-memory}

\method{} can also serve as the retrieval component of a
memory system. On LoCoMo, each reranker ordered the same 50
Qwen3-Embedding-8B candidates, and only the top-3 chunks were passed to
a generator LLM for question answering; Appendix~\ref{app:memory-qa}
describes the setup and reports full results. Reranking raises overall
F1 over the first-stage order by 16.48--18.79 points, whereas switching
among GPT-4o-mini, GPT-5-mini, and GPT-6-luna as the generator changes
it by at most 2.07 points. With GPT-4o-mini, all three \method{}
variants achieve higher overall F1 than every memory framework reported
by \citet{li2026queryfocusedmemoryawarererankerlong}.

\subsection{Efficiency}
\label{sec:efficiency}

\begin{figure}[t]
\centering
\includegraphics[width=\columnwidth]{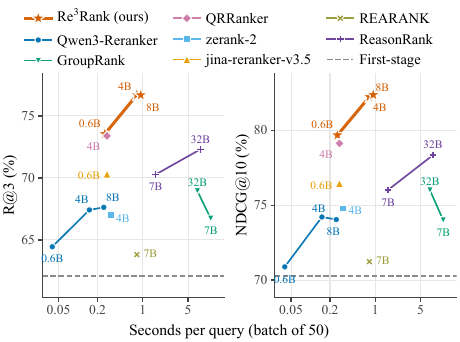}
\caption{Macro-9 R@3 (left) and NDCG@10 (right) against reranking
time per query. For each test set, 50 sampled queries were submitted as
one batch, and the wall-clock time until all were scored was divided by
50 and averaged over the nine test sets.}
\label{fig:metric-vs-latency}
\end{figure}

Figure~\ref{fig:metric-vs-latency} compares accuracy with reranking
time per query under batched inference. \method{} (0.6B) is slightly
faster than QRRanker and jina-reranker-v3.5 (0.26 vs.\ 0.28\,s) while
achieving higher accuracy, and the 4B and 8B
variants add 3.3 R@3 points over QRRanker at 0.81 and 0.93\,s per
query, respectively.
Truncating the reasoning trace at inference further trades accuracy for
speed without retraining (Appendix~\ref{app:trace-budget}). 

\subsection{Generality}
\label{sec:generality}

\begin{table}[t]
\centering
\footnotesize
\setlength{\tabcolsep}{2pt}
\begin{tabular*}{\columnwidth}{@{\extracolsep{\fill}}lcccc@{}}
\toprule
\textbf{Backbone} & \textbf{R@1} & \textbf{R@3} & \textbf{R@5} & \textbf{N@10}\\
\midrule
Qwen3-0.6B & 42.34 & 73.55 & 81.43 & 79.67\\
Qwen3-4B & 44.20 & 76.70 & 84.01 & 82.23\\
Qwen3-8B & 44.12 & 76.68 & \textbf{84.36} & 82.36\\
\midrule
Gemma-4-E2B & 44.00 & 75.44 & 82.99 & 81.47\\
Gemma-4-E4B & \textbf{44.51} & \textbf{77.14} & 84.06 & \textbf{82.54}\\
\bottomrule
\end{tabular*}
\caption{Macro-9 results (\%) of \method{} with Qwen3 and Gemma~4
backbones, all trained with $\mathcal{L}_{\mathrm{rank}}$. Bold marks
the best result in each column.}
\label{tab:gemma}
\end{table}

\method{} is not tied to a particular model family. To verify this, we
applied the same training procedure to Gemma-4-E2B and Gemma-4-E4B
\citep{gemmateam2026gemma4}, selecting their RRHeads as in
Section~\ref{sec:rrhead-selection}. As Table~\ref{tab:gemma} shows,
both Gemma variants reach Macro-9 accuracy comparable to the Qwen3
variants and outperform every baseline on all four metrics. The mixed
objective also improves over $\mathcal{L}_{\mathrm{CE}}$ on both Gemma
backbones, as it does on Qwen3-4B (Appendix~\ref{app:gemma}).

\FloatBarrier
\section{Conclusion}
\label{sec:conclusion}

We presented \method{}, a listwise reranker built on
reasoning-conditioned retrieval heads. \method{} uses a generated reasoning trace as the attention
query of a few selected RRHeads and reads relevance as continuous
attention scores, without verbalizing a permutation. 
A straight-through
estimator carries the ranking loss through the discrete reasoning
trace, so that the generator and the scorer can be trained jointly.
On nine
benchmarks, \method{} outperforms existing rerankers across model
scales, with its smallest 0.6B variant already surpassing much larger
baselines. 
Its design transfers across backbones, and its gains in
downstream QA suggest
its potential as the retrieval component of agentic memory systems.

\section*{Limitations}

The generator of \method{} is optimized only through the ranking loss,
so its reasoning traces carry no guarantee of linguistic quality or
readability to humans. In addition, all
experiments rerank a fixed pool of the top 50 candidates returned by a
single first-stage retriever, Qwen3-Embedding-8B. Owing to constraints
on computational resources and experimental scale, the correctness of
the traces and the behavior of \method{} with other retrievers or with
larger candidate pools, which may exceed the context window, remain to be explored.

\bibliography{custom}

\appendix
\raggedbottom
\renewcommand{\dbltopfraction}{0.95}

\section{RRHead Selection for Qwen3-4B}
\label{app:rrhead-selection}

This appendix details the head scan behind the Qwen3-4B RRHeads
$\{(23,10),(21,8)\}$ reported in Section~\ref{sec:rrhead-selection}
and examines why two heads suffice.
All head indices are zero-based (layer, head) pairs.

\paragraph{Seed set.}
$\mathcal{D}_{\mathrm{seed}}$ contained 400 instances each from the
MuSiQue and NarrativeQA portions of the training mixture
(Section~\ref{sec:training-data}). Instances were sampled with a fixed
random seed from questions whose 50-candidate pool contains at least
one relevant candidate, giving 2.31 relevant candidates per question on
average. As training questions, they do not overlap with the evaluation
questions in Section~\ref{sec:benchmarks}. On
$\mathcal{D}_{\mathrm{seed}}$, the first-stage order reaches 51.78 R@3
and 64.14 NDCG@10.

\paragraph{Scan.}
A single forward pass over $\mathbf{x}\mathbin{\|}\mathbf{u}$ yielded,
for every head, the attention weights $A^{(l,h)}_{t,j}$
(Section~\ref{sec:qrhead}) from the trace positions
$\mathcal{Q}_u$, which were converted into candidate masses by
Eq.~\ref{eq:head-mass}. Candidate spans $\mathcal{I}_i$ covered only the
candidate text, excluding index markers and separators. Qwen3-4B uses
grouped-query attention with 32 query heads and 8 key--value heads per
layer; $h$ indexes query heads.

\paragraph{Per-head statistics.}
Besides the gold mass $\gamma^{(l,h)}$ in Eq.~\ref{eq:head-selection},
each head was characterized by its gold share and its zero-shot ranking
quality. The \emph{gold share} averages
$\sum_{i\in\mathcal{P}}m_i^{(l,h)}/\sum_{i=1}^{C}m_i^{(l,h)}$ over
$\mathcal{D}_{\mathrm{seed}}$ and measures how concentrated the
candidate-directed attention is on relevant candidates. Zero-shot
Recall@3 (Eq.~\ref{eq:reranking-objective}
with $k=3$) and binary NDCG@10 were obtained by ranking candidates
directly by $m_i^{(l,h)}$.

\begin{figure}[t]
\centering
\includegraphics[width=\columnwidth]{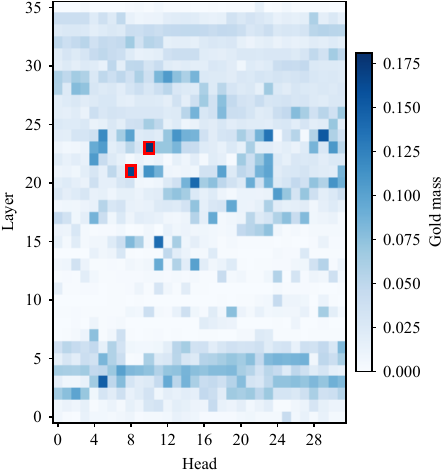}
\caption{Gold mass $\gamma^{(l,h)}$ of all $36\times32$ Qwen3-4B heads
on $\mathcal{D}_{\mathrm{seed}}$. Red boxes mark the selected RRHeads
$(23,10)$ and $(21,8)$.}
\label{fig:rrhead-heatmap}
\end{figure}

\paragraph{Selection.}
$\mathcal{H}_{\mathrm{RR}}$ consisted of the two heads with the highest
gold mass, $(23,10)$ and $(21,8)$.
Figure~\ref{fig:rrhead-heatmap} shows $\gamma^{(l,h)}$ for all 1,152
heads, and Table~\ref{tab:rrhead-top12} lists the twelve heads with the
highest gold mass.

\begin{table}[t]
\centering
\small
\begin{tabular*}{\columnwidth}{@{\extracolsep{\fill}}rlcccc@{}}
\toprule
\textbf{Rank} & \textbf{Head} &
\begin{tabular}[b]{@{}c@{}}\textbf{Gold}\\\textbf{mass}\end{tabular} &
\begin{tabular}[b]{@{}c@{}}\textbf{Gold}\\\textbf{share}\end{tabular} &
\textbf{R@3} & \textbf{N@10}\\
\midrule
1  & (23,10) & \textbf{0.1813} & 0.395 & 59.42 & 70.48\\
2  & (21,8)  & \textbf{0.1669} & 0.332 & 59.24 & 69.78\\
3  & (24,29) & 0.1526 & 0.336 & 59.14 & 70.01\\
4  & (3,5)   & 0.1487 & 0.178 & 52.32 & 60.91\\
5  & (20,15) & 0.1412 & 0.330 & 58.73 & 68.75\\
6  & (15,11) & 0.1386 & 0.297 & 59.42 & 70.14\\
7  & (24,23) & 0.1328 & 0.325 & 55.30 & 66.44\\
8  & (24,5)  & 0.1181 & 0.323 & 53.75 & 66.29\\
9  & (21,10) & 0.1120 & 0.358 & 58.92 & 69.10\\
10 & (20,30) & 0.1113 & 0.297 & 54.59 & 66.09\\
11 & (24,13) & 0.1100 & 0.318 & 58.35 & 68.32\\
12 & (23,30) & 0.1079 & 0.341 & 57.97 & 68.72\\
\bottomrule
\end{tabular*}
\caption{Top-12 Qwen3-4B heads on $\mathcal{D}_{\mathrm{seed}}$, sorted
by gold mass. R@3 and N@10 are zero-shot single-head Recall@3 and
NDCG@10 (\%). Bold marks the two highest gold masses, whose heads form
$\mathcal{H}_{\mathrm{RR}}$.}
\label{tab:rrhead-top12}
\end{table}

\paragraph{Number of heads.}
Summing more heads does not reliably improve zero-shot ranking.
Table~\ref{tab:rrhead-count} sums the masses of the top-$n$ heads by
gold mass, and differences are tested with a paired bootstrap over seed
instances (2,000 resamples, stratified by dataset). The top two reach
59.96 R@3 against 59.42 for $(23,10)$ alone, a difference that is not
significant (95\% interval $[-0.34,+1.39]$). The top four reach the
highest R@3, 60.93, but their gain over the top two is not significant
either ($[-0.04,+2.05]$) and comes with a slightly lower NDCG@10. Beyond
four heads, both metrics decline: the top 16 heads, the number used by
QRHead \citep{zhang-etal-2025-query-focused}, fall 1.40 R@3 points
below the top two ($[-2.41,-0.46]$), and the top 64 fall 4.38 points
below ($[-6.05,-2.73]$). Two heads therefore perform on par with the
best number of heads, whereas adding many more heads hurts ranking.

\begin{table}[t]
\centering
\small
\setlength{\tabcolsep}{2pt}
\begin{tabular*}{\columnwidth}{@{\extracolsep{\fill}}lccccccc@{}}
\toprule
$n$ & 1 & 2 & 4 & 8 & 16 & 32 & 64\\
\midrule
R@3  & 59.42 & 59.96 & \textbf{60.93} & 59.58 & 58.56 & 58.14 & 55.58\\
N@10 & 70.48 & \textbf{70.71} & 70.55 & 70.42 & 69.35 & 67.92 & 63.99\\
\bottomrule
\end{tabular*}
\caption{Zero-shot R@3 and NDCG@10 (\%) of Qwen3-4B on
$\mathcal{D}_{\mathrm{seed}}$ when the masses of the top-$n$ heads by
gold mass are summed. $n=1$ is $(23,10)$ alone, and $n=2$ gives
$\mathcal{H}_{\mathrm{RR}}$. Bold marks the best value in each row.}
\label{tab:rrhead-count}
\end{table}

\paragraph{Redundancy among heads.}
Additional heads contribute little because the strongest heads largely
duplicate one another. For each seed instance and each of its relevant
candidates, a head \emph{hits} the candidate if it ranks the candidate
among the top three of the 50 candidates, and \emph{misses} it
otherwise. Comparing which relevant candidates two heads hit shows
whether they locate the same evidence, and
Table~\ref{tab:rrhead-overlap} compares every other head with
$(23,10)$ in this way; $(23,10)$ itself hits 55.1\% of the 1,848
relevant candidates in $\mathcal{D}_{\mathrm{seed}}$.
Whatever their strength, other heads hit only 8.1--9.2\% of the
relevant candidates that $(23,10)$ misses, whereas the first-stage
ranking, a different relevance signal, hits 23.6\%; a head as accurate
as $(23,10)$ but with independent errors would hit about 55\%.
Conversely, the weaker a head is, the more of its misses $(23,10)$
hits, rising from 10.9\% for the 15 strongest heads to 52.9\% for the
weakest 896, close to the 55.2\% for a random ranking. Strong heads
therefore miss the same candidates as $(23,10)$, and most hits of weak
heads are also hits of $(23,10)$.

\begin{table}[t]
\centering
\small
\setlength{\tabcolsep}{2pt}
\begin{tabular*}{\columnwidth}{@{\extracolsep{\fill}}lrccc@{}}
\toprule
\textbf{Ranking} & \textbf{\#Heads} & \textbf{Shared} & \textbf{Gain} &
\textbf{Covered}\\
\midrule
Rank 1--16 (excl.\,$(23,10)$) & 15  & 92.6 & 9.0 & 10.9\\
Rank 17--64     & 48  & 91.9 & 9.2 & 18.8\\
Rank 65--256    & 192 & 88.0 & 9.0 & 38.2\\
Rank 257--1152  & 896 & 74.1 & 8.1 & 52.9\\
\midrule
Random ranking & -- & 54.5 & 6.2 & 55.2\\
First-stage    & -- & 77.6 & 23.6 & 35.0\\
\bottomrule
\end{tabular*}
\caption{Overlap of top-3 hits between $(23,10)$ and other rankings on
$\mathcal{D}_{\mathrm{seed}}$ (\%). Heads other than $(23,10)$ are
grouped by their rank in single-head R@3 among all 1,152 heads, and
each group reports the median over its heads; the random ranking
averages 20 random orders. Shared is the share of a ranking's hits that
$(23,10)$ also hits, Gain is the share of the misses of $(23,10)$ that
the ranking hits, and Covered is the share of the ranking's misses that
$(23,10)$ hits.}
\label{tab:rrhead-overlap}
\end{table}

\section{Full Reranking Results}
\label{app:full-results}

\begin{table*}[t]
\centering
\small
\setlength{\tabcolsep}{2.5pt}
\begin{tabular}{lcccccccccccc}
\toprule
\multicolumn{13}{c}{\textbf{Wikipedia multi-hop}}\\
\midrule
\textbf{Method}
& \multicolumn{4}{c}{\textbf{HotpotQA}}
& \multicolumn{4}{c}{\textbf{2WikiMultihopQA}}
& \multicolumn{4}{c}{\textbf{MuSiQue}}\\
\cmidrule(lr){2-5}\cmidrule(lr){6-9}\cmidrule(lr){10-13}
 & \textbf{R@1} & \textbf{R@3} & \textbf{R@5} & \textbf{N@10} & \textbf{R@1} & \textbf{R@3} & \textbf{R@5} & \textbf{N@10} & \textbf{R@1} & \textbf{R@3} & \textbf{R@5} & \textbf{N@10}\\
\midrule
First-stage & 46.75 & 83.08 & 90.29 & 89.40 & 53.42 & 86.08 & 89.05 & 91.38 & 41.55 & 64.62 & 73.83 & 76.42\\
\midrule
QRRanker (4B) & 49.05 & 96.40 & 98.80 & 96.62 & 54.22 & 98.38 & 99.58 & 98.87 & 44.58 & 81.44 & 89.67 & 88.84\\
Qwen3-Reranker (0.6B) & 48.10 & 86.49 & 91.84 & 91.23 & 54.20 & 84.78 & 87.66 & 90.38 & 41.29 & 66.02 & 73.69 & 74.91\\
Qwen3-Reranker (4B) & 49.10 & 91.09 & 96.05 & 94.49 & 53.85 & 87.31 & 90.19 & 92.28 & 42.25 & 69.92 & 78.64 & 79.71\\
Qwen3-Reranker (8B) & 48.90 & 91.94 & 95.65 & 94.41 & 51.45 & 88.33 & 90.99 & 91.92 & 39.71 & 69.38 & 79.25 & 78.71\\
zerank-2 (4B) & 45.70 & 84.78 & 92.64 & 90.36 & 44.67 & 81.73 & 87.62 & 86.58 & 34.94 & 58.96 & 70.71 & 70.90\\
jina-reranker-v3.5 (0.6B) & 49.25 & 95.95 & 98.30 & 96.57 & \textbf{54.90} & 94.51 & 97.12 & 97.50 & 45.28 & 76.02 & 83.01 & 84.71\\
GroupRank (7B) & 44.69 & 83.18 & 92.44 & 89.13 & 47.38 & 84.00 & 90.30 & 88.23 & 38.93 & 67.11 & 77.20 & 77.22\\
GroupRank (32B) & 45.45 & 86.64 & 93.69 & 90.65 & 45.52 & 83.00 & 89.96 & 87.21 & 37.77 & 66.78 & 78.38 & 77.75\\
REARANK (7B) & 46.75 & 88.54 & 94.84 & 92.18 & 47.09 & 87.16 & 92.27 & 90.46 & 34.77 & 63.88 & 75.67 & 74.39\\
ReasonRank (7B) & 48.60 & 94.24 & 96.75 & 95.29 & 51.77 & 93.83 & 95.08 & 94.68 & 38.18 & 66.63 & 74.24 & 74.87\\
ReasonRank (32B) & 49.10 & 94.84 & 97.40 & 95.77 & 53.20 & 96.41 & 97.24 & 97.05 & 38.97 & 68.57 & 76.67 & 77.49\\
\midrule
\method{} (0.6B) & 48.85 & 95.85 & 98.45 & 96.38 & 54.17 & 98.17 & \underline{99.90} & 98.90 & 44.94 & 79.28 & 88.15 & 88.36\\
\method{} (4B) & \textbf{50.35} & \underline{97.60} & \underline{99.25} & \underline{97.95} & \underline{54.57} & \underline{99.13} & \textbf{99.95} & \underline{99.50} & \underline{45.71} & \underline{84.35} & \underline{91.43} & \underline{90.73}\\
\method{} (8B) & \underline{50.00} & \textbf{98.15} & \textbf{99.30} & \textbf{98.03} & 54.48 & \textbf{99.15} & \textbf{99.95} & \textbf{99.52} & \textbf{47.00} & \textbf{86.02} & \textbf{93.10} & \textbf{92.57}\\
\midrule
$\Delta$ vs.\ best baseline & \deltapos{+1.10} & \deltapos{+1.75} & \deltapos{+0.50} & \deltapos{+1.41} & \deltaneg{$-$0.33} & \deltapos{+0.77} & \deltapos{+0.37} & \deltapos{+0.65} & \deltapos{+1.72} & \deltapos{+4.58} & \deltapos{+3.43} & \deltapos{+3.73}\\
Significance & *** & *** & n.s. & *** & n.s. & ** & ** & *** & ** & *** & *** & ***\\
\bottomrule
\end{tabular}
\caption{Reranking results on the Wikipedia multi-hop datasets. R@$k$ is conditional Recall@$k$ and N@10 is NDCG@10 (\%). First-stage is the Qwen3-Embedding-8B ranking before reranking. \method{} was trained with the mixed ranking loss in Eq.~\ref{eq:mixed-ranking-loss}. Bold and underline mark the best and second-best reranker for each metric. The $\Delta$ row gives the difference between the best \method{} result and the best baseline result in each column, with gains in green and losses in red. The significance row reports two-sided paired randomization tests on per-question scores (100,000 sign flips) for the same pair: *** $p<0.001$, ** $p<0.01$, * $p<0.05$, and n.s.\ not significant.}
\label{tab:main-wiki}
\end{table*}

\begin{table*}[t]
\centering
\small
\setlength{\tabcolsep}{5pt}
\begin{tabular}{lcccccccc}
\toprule
\multicolumn{9}{c}{\textbf{Open-domain factoid}}\\
\midrule
\textbf{Method}
& \multicolumn{4}{c}{\textbf{NQ}}
& \multicolumn{4}{c}{\textbf{PopQA}}\\
\cmidrule(lr){2-5}\cmidrule(lr){6-9}
 & \textbf{R@1} & \textbf{R@3} & \textbf{R@5} & \textbf{N@10} & \textbf{R@1} & \textbf{R@3} & \textbf{R@5} & \textbf{N@10}\\
\midrule
First-stage & 25.80 & 54.57 & 72.45 & 81.49 & 54.65 & 89.90 & 91.90 & 79.66\\
\midrule
QRRanker (4B) & 23.76 & 55.18 & 73.82 & 79.30 & 54.50 & 90.70 & 96.70 & 81.84\\
Qwen3-Reranker (0.6B) & 23.14 & 50.79 & 69.45 & 78.29 & \textbf{57.55} & 89.55 & 91.55 & 79.92\\
Qwen3-Reranker (4B) & 26.76 & 56.79 & 75.00 & 83.02 & 53.60 & 89.75 & 93.00 & 80.05\\
Qwen3-Reranker (8B) & 27.32 & 58.23 & 76.33 & 84.09 & 49.95 & 88.45 & 94.05 & 78.55\\
zerank-2 (4B) & 28.32 & 58.32 & 77.42 & \underline{84.56} & 56.80 & 89.95 & 92.65 & 82.06\\
jina-reranker-v3.5 (0.6B) & 28.50 & 57.71 & 74.69 & 82.94 & \underline{57.50} & 91.80 & 96.20 & \underline{83.48}\\
GroupRank (7B) & 27.55 & 58.15 & 76.37 & 83.63 & 51.55 & 87.25 & 91.95 & 78.29\\
GroupRank (32B) & 28.63 & \underline{60.17} & 77.90 & \textbf{84.75} & 51.20 & 88.20 & 93.40 & 78.99\\
REARANK (7B) & 24.70 & 54.19 & 71.05 & 79.92 & 55.80 & 90.45 & 91.90 & 80.21\\
ReasonRank (7B) & \textbf{28.85} & 59.40 & 75.30 & 83.06 & 55.90 & 91.10 & 94.40 & 81.59\\
ReasonRank (32B) & \underline{28.64} & 59.99 & 76.87 & 84.23 & 56.80 & 90.90 & 94.95 & 81.91\\
\midrule
\method{} (0.6B) & 23.02 & 55.25 & 75.78 & 80.92 & 57.15 & \underline{94.50} & \underline{97.80} & \textbf{84.14}\\
\method{} (4B) & 27.44 & \textbf{60.50} & \textbf{78.73} & \textbf{84.75} & 54.95 & \textbf{95.40} & \textbf{98.75} & 83.30\\
\method{} (8B) & 26.69 & 60.06 & \underline{78.40} & 84.07 & 52.50 & 92.80 & 97.60 & 81.60\\
\midrule
$\Delta$ vs.\ best baseline & \deltaneg{$-$1.41} & \deltapos{+0.33} & \deltapos{+0.83} & 0.00 & \deltaneg{$-$0.40} & \deltapos{+3.60} & \deltapos{+2.05} & \deltapos{+0.66}\\
Significance & n.s. & n.s. & ** & n.s. & n.s. & *** & *** & n.s.\\
\bottomrule
\end{tabular}
\caption{Reranking results on the open-domain factoid datasets. Metrics and markers follow Table~\ref{tab:main-wiki}.}
\label{tab:main-open}
\end{table*}

\begin{table*}[t]
\centering
\small
\setlength{\tabcolsep}{5pt}
\begin{tabular}{lcccccccc}
\toprule
\multicolumn{9}{c}{\textbf{Long narrative}}\\
\midrule
\textbf{Method}
& \multicolumn{4}{c}{\textbf{NarrativeQA}}
& \multicolumn{4}{c}{\textbf{DetectiveQA}}\\
\cmidrule(lr){2-5}\cmidrule(lr){6-9}
 & \textbf{R@1} & \textbf{R@3} & \textbf{R@5} & \textbf{N@10} & \textbf{R@1} & \textbf{R@3} & \textbf{R@5} & \textbf{N@10}\\
\midrule
First-stage & 16.27 & 38.59 & 52.65 & 52.91 & 9.38 & 19.87 & 27.17 & 29.65\\
\midrule
QRRanker (4B) & 20.88 & 42.30 & 56.01 & 55.95 & 11.33 & 24.76 & 33.50 & 34.35\\
Qwen3-Reranker (0.6B) & 15.85 & 30.09 & 39.33 & 40.59 & 6.75 & 17.94 & 25.59 & 25.96\\
Qwen3-Reranker (4B) & 19.04 & 36.81 & 47.81 & 48.65 & 10.15 & 19.53 & 29.05 & 29.85\\
Qwen3-Reranker (8B) & 20.78 & 39.10 & 50.51 & 51.50 & 9.71 & 20.17 & 28.32 & 30.05\\
zerank-2 (4B) & 20.95 & 40.13 & 52.31 & 53.12 & 10.89 & 20.73 & 28.00 & 30.66\\
jina-reranker-v3.5 (0.6B) & 16.51 & 34.25 & 45.03 & 45.93 & 8.07 & 20.04 & 27.10 & 28.28\\
GroupRank (7B) & 22.55 & 42.43 & 54.11 & 54.78 & 8.17 & 20.02 & 29.19 & 29.86\\
GroupRank (32B) & 24.98 & 46.54 & 60.45 & 60.40 & 9.25 & 22.17 & 30.29 & 31.50\\
REARANK (7B) & 15.84 & 33.79 & 46.25 & 47.53 & 7.84 & 18.45 & 27.97 & 28.22\\
ReasonRank (7B) & 20.57 & 43.97 & 55.91 & 55.23 & 9.41 & 23.68 & 32.18 & 33.32\\
ReasonRank (32B) & 25.76 & 46.87 & 57.43 & 58.99 & 9.72 & 23.84 & 33.75 & 34.26\\
\midrule
\method{} (0.6B) & 22.67 & 47.23 & 60.36 & 60.71 & 10.09 & 22.46 & 31.77 & 32.97\\
\method{} (4B) & \underline{26.80} & \underline{53.19} & \underline{67.50} & \underline{66.95} & \textbf{12.36} & \textbf{26.11} & \underline{35.55} & \textbf{37.12}\\
\method{} (8B) & \textbf{28.07} & \textbf{54.60} & \textbf{68.65} & \textbf{68.52} & \underline{11.57} & \underline{24.85} & \textbf{36.67} & \underline{36.38}\\
\midrule
$\Delta$ vs.\ best baseline & \deltapos{+2.31} & \deltapos{+7.73} & \deltapos{+8.20} & \deltapos{+8.12} & \deltapos{+1.03} & \deltapos{+1.35} & \deltapos{+2.92} & \deltapos{+2.77}\\
Significance & *** & *** & *** & *** & n.s. & n.s. & * & **\\
\bottomrule
\end{tabular}
\caption{Reranking results on the long-narrative datasets. Metrics and markers follow Table~\ref{tab:main-wiki}.}
\label{tab:main-narrative}
\end{table*}

\begin{table*}[t]
\centering
\small
\setlength{\tabcolsep}{2.5pt}
\begin{tabular}{lcccccccccccc}
\toprule
\multicolumn{13}{c}{\textbf{Long dialogue and overall}}\\
\midrule
\textbf{Method}
& \multicolumn{4}{c}{\textbf{LoCoMo}}
& \multicolumn{4}{c}{\textbf{LongMemEval-S}}
& \multicolumn{4}{c}{\textbf{Macro-9}}\\
\cmidrule(lr){2-5}\cmidrule(lr){6-9}\cmidrule(lr){10-13}
 & \textbf{R@1} & \textbf{R@3} & \textbf{R@5} & \textbf{N@10} & \textbf{R@1} & \textbf{R@3} & \textbf{R@5} & \textbf{N@10} & \textbf{R@1} & \textbf{R@3} & \textbf{R@5} & \textbf{N@10}\\
\midrule
First-stage & 33.88 & 53.66 & 62.00 & 55.08 & 40.90 & 68.48 & 79.10 & 76.47 & 35.84 & 62.09 & 70.94 & 70.27\\
\midrule
QRRanker (4B) & 71.73 & 88.43 & 92.01 & 88.56 & 50.75 & 82.94 & 90.31 & 87.72 & 42.31 & 73.39 & 81.16 & 79.12\\
Qwen3-Reranker (0.6B) & 61.98 & 82.97 & 88.12 & 81.62 & 44.26 & 71.40 & 78.29 & 74.98 & 39.24 & 64.45 & 71.72 & 70.88\\
Qwen3-Reranker (4B) & 63.77 & 82.65 & 87.63 & 82.27 & 44.59 & 72.89 & 80.77 & 77.56 & 40.35 & 67.42 & 75.35 & 74.21\\
Qwen3-Reranker (8B) & 63.28 & 81.94 & 86.99 & 81.67 & 45.11 & 71.14 & 77.43 & 75.46 & 39.58 & 67.63 & 75.50 & 74.04\\
zerank-2 (4B) & 72.16 & \underline{89.13} & \underline{93.10} & \underline{89.17} & 48.22 & 79.29 & 89.27 & 85.53 & 40.29 & 67.00 & 75.97 & 74.77\\
jina-reranker-v3.5 (0.6B) & 72.16 & 87.07 & 90.70 & 87.90 & 47.60 & 75.44 & 82.90 & 80.77 & 42.20 & 70.31 & 77.23 & 76.45\\
GroupRank (7B) & 66.09 & 86.51 & 91.07 & 85.56 & 43.16 & 71.91 & 82.24 & 79.33 & 38.90 & 66.73 & 76.10 & 74.00\\
GroupRank (32B) & 70.81 & 88.59 & 92.70 & 88.35 & 46.65 & 78.36 & 88.87 & 84.43 & 40.03 & 68.94 & 78.40 & 76.00\\
REARANK (7B) & 41.81 & 59.70 & 69.60 & 63.36 & 48.01 & 78.24 & 87.66 & 84.84 & 35.85 & 63.82 & 73.02 & 71.23\\
ReasonRank (7B) & 68.51 & 85.83 & 89.65 & 85.41 & 46.84 & 73.84 & 80.92 & 80.71 & 40.96 & 70.28 & 77.16 & 76.02\\
ReasonRank (32B) & 71.69 & 88.56 & 92.29 & 88.57 & 50.22 & 80.78 & 88.76 & 86.77 & 42.68 & 72.31 & 79.48 & 78.34\\
\midrule
\method{} (0.6B) & 68.07 & 85.40 & 89.66 & 85.55 & 52.12 & 83.79 & 90.96 & 89.10 & 42.34 & 73.55 & 81.43 & 79.67\\
\method{} (4B) & \underline{72.53} & 88.13 & 92.26 & 88.96 & \textbf{53.07} & \textbf{85.86} & \textbf{92.67} & \textbf{90.81} & \textbf{44.20} & \textbf{76.70} & \underline{84.01} & \underline{82.23}\\
\method{} (8B) & \textbf{74.11} & \textbf{89.50} & \textbf{93.30} & \textbf{90.16} & \underline{52.66} & \underline{85.03} & \underline{92.29} & \underline{90.43} & \underline{44.12} & \underline{76.68} & \textbf{84.36} & \textbf{82.36}\\
\midrule
$\Delta$ vs.\ best baseline & \deltapos{+1.95} & \deltapos{+0.37} & \deltapos{+0.20} & \deltapos{+0.99} & \deltapos{+2.32} & \deltapos{+2.92} & \deltapos{+2.36} & \deltapos{+3.09} & \deltapos{+1.52} & \deltapos{+3.31} & \deltapos{+3.20} & \deltapos{+3.24}\\
Significance & * & n.s. & n.s. & * & n.s. & *** & ** & *** & ** & *** & *** & ***\\
\bottomrule
\end{tabular}
\caption{Reranking results on the long-dialogue datasets and the macro average over all nine datasets (Macro-9). Metrics and markers follow Table~\ref{tab:main-wiki}.}
\label{tab:main-dialogue}
\end{table*}

\begin{figure*}[t]
  \centering
  \includegraphics[width=\textwidth]{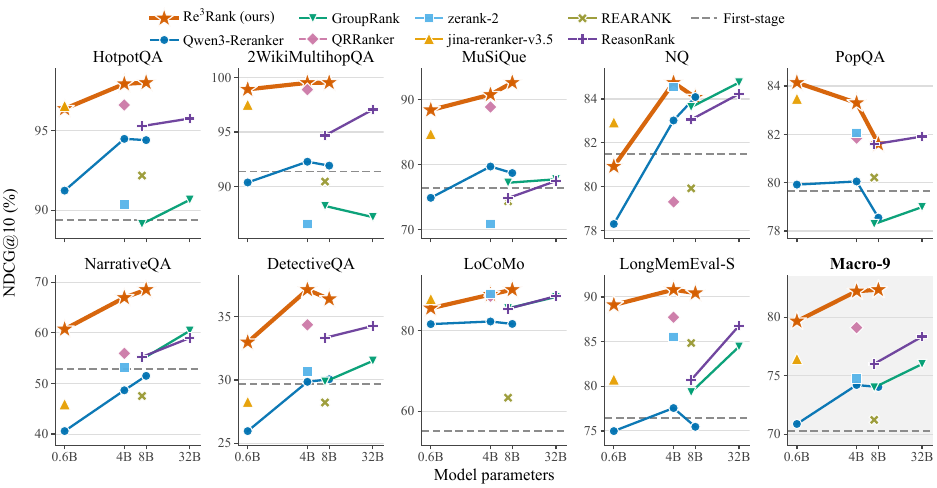}
  \caption{NDCG@10 (\%) of rerankers across model sizes on the nine
  benchmarks and their macro average (Macro-9). Lines and markers follow
  Figure~\ref{fig:r3-vs-size}.}
  \label{fig:ndcg10-vs-size}
\end{figure*}

Tables~\ref{tab:main-wiki}--\ref{tab:main-dialogue} report conditional
R@1, R@3, R@5, and NDCG@10 of each reranker on each benchmark, together
with the gap between the best \method{} variant and the best baseline
and its statistical significance. Figure~\ref{fig:ndcg10-vs-size}
plots NDCG@10 across model sizes in the format of
Figure~\ref{fig:r3-vs-size}.

\paragraph{Evaluation protocol.}
Every reranker received the same 50 first-stage candidates per question. For a question, let
$n_{\mathrm{in}}$ be the number of relevant chunks among its candidates.
Questions with $n_{\mathrm{in}}=0$ were excluded, so all metrics are
conditional on the candidate pool. Let $y_{(t)}\in\{0,1\}$ be the
relevance label of the candidate ranked at position $t$ by a reranker.
Recall@$k$, which matches Eq.~\ref{eq:reranking-objective}, and
normalized discounted cumulative gain (NDCG@10) with binary relevance
\citep{jarvelin2002cumulated} are
\begin{align*}
\mathrm{R@}k
&=\frac{1}{n_{\mathrm{in}}}\sum_{t=1}^{k}y_{(t)},\\
\mathrm{NDCG@}10
&=\frac{\sum_{t=1}^{10}y_{(t)}/\log_2(t+1)}
{\sum_{t=1}^{\min(10,\,n_{\mathrm{in}})}1/\log_2(t+1)},
\end{align*}
where the numerator is the discounted cumulative gain (DCG) of the
reranked list and the denominator is the DCG of an ideal ranking that
places all in-pool relevant chunks first. Each metric was averaged over questions
within a dataset, and Macro-9 is the unweighted mean over the nine
datasets.

\paragraph{Baseline inference.}
QRRanker was evaluated with its official release and evaluation
code.\footnote{\url{https://huggingface.co/MindscapeRAG/QRRanker}}
Qwen3-Reranker followed its
official prompt and scored each question--candidate pair by the
probability of ``yes'' against ``no'' at the final position.\footnote{\url{https://huggingface.co/Qwen/Qwen3-Reranker-0.6B},
\url{https://huggingface.co/Qwen/Qwen3-Reranker-4B}, and
\url{https://huggingface.co/Qwen/Qwen3-Reranker-8B}}
zerank-2 was scored by its raw ``Yes''
logit.\footnote{\url{https://huggingface.co/zeroentropy/zerank-2}}
jina-reranker-v3.5 scored all 50 candidates in a single forward pass
through its listwise
interface.\footnote{\url{https://huggingface.co/jinaai/jina-reranker-v3.5}}
GroupRank scored candidates in groups of 10 and averaged the scores
over two shuffled
passes.\footnote{\url{https://huggingface.co/AQ-MedAI/Diver-GroupRank-7B}
and \url{https://huggingface.co/AQ-MedAI/Diver-GroupRank-32B}}
REARANK and ReasonRank followed their official inference
settings.\footnote{\url{https://huggingface.co/le723z/Rearank-7B},
\url{https://huggingface.co/liuwenhan/reasonrank-7B}, and
\url{https://huggingface.co/liuwenhan/reasonrank-32B}}

\paragraph{Significance testing.}
The significance row tests whether the gap in the $\Delta$ row is
reliable, comparing the best \method{} variant with the best baseline
on the same questions. A two-sided paired randomization test was applied
to their per-question scores: the sign of each question's score
difference was flipped at random 100,000 times, and the $p$-value is the
fraction of these flips whose mean difference is at least as large in
magnitude as the observed one. The test does not assume normally
distributed differences, which suits the discrete per-question values
of Recall@$k$.
Markers denote *** $p<0.001$, ** $p<0.01$,
* $p<0.05$, and n.s.\ $p\ge0.05$, without correction for multiple
comparisons.

\paragraph{Significance results.}
Of the 36 dataset-level columns, 24 show a significant difference, 15
at $p<0.001$, 6 at $p<0.01$, and 3 at $p<0.05$, and all four Macro-9
columns are significant. Every significant difference favors \method{}.

\section{Implementation Details}
\label{app:implementation}

\subsection{Training Hyperparameters}
\label{app:hyperparams}

Training used one epoch with a global batch size of 96. The generator
and scorer were optimized with AdamW using a constant
learning rate of $10^{-5}$.
Reasoning traces were sampled with temperature 1, top-$p=1$, no
top-$k$ truncation, and a maximum length of 1,024 tokens. The
straight-through relaxation used $\tau_{\mathrm{ST}}=1$.

\subsection{Corpora and Chunking}
\label{app:corpora}

\begin{table*}[t]
\centering
\small
\setlength{\tabcolsep}{4pt}
\begin{tabular}{llrlrrrr}
\toprule
& & & & & & \multicolumn{2}{c}{\textbf{Gold / question}}\\
\cmidrule(lr){7-8}
\textbf{Dataset} & \textbf{Regime} & \textbf{\#Questions} &
\textbf{Retrieval scope} & \textbf{\#Chunks} & \textbf{Tok./chunk} &
\textbf{All} & \textbf{In-pool}\\
\midrule
HotpotQA        & Wiki multi-hop & 1,000 & Shared corpus         & 9,811   & 138 & 2.00 & 1.96\\
2WikiMultihopQA & Wiki multi-hop & 1,000 & Shared corpus         & 6,119   & 118 & 2.47 & 1.98\\
MuSiQue         & Wiki multi-hop & 1,000 & Shared corpus         & 11,656  & 121 & 2.65 & 2.21\\
NQ              & Open-domain    & 1,000 & Shared corpus         & 9,633   & 148 & 3.79 & 3.79\\
PopQA           & Open-domain    & 1,000 & Shared corpus         & 8,676   & 146 & 2.00 & 1.17\\
NarrativeQA     & Long narrative & 1,902 & Per book (65)         & 59,043  & 200 & 2.93 & 2.39\\
DetectiveQA     & Long narrative &   600 & Per novel (62)        & 31,807  & 173 & 5.57 & 2.31\\
LoCoMo          & Long dialogue  & 1,531 & Per conversation (10) & 849     & 261 & 1.42 & 1.39\\
LongMemEval-S   & Long dialogue  &   466 & Per chat history (466) & 114,078 & 255 & 2.45 & 2.31\\
\midrule
Total           &                & 9,499 &                       & 251,672 &     &      &     \\
\bottomrule
\end{tabular}
\caption{Evaluation benchmarks. For each question, Qwen3-Embedding-8B
served as the first-stage retriever and returned the top-50 candidates
from the retrieval scope; parentheses give the number of scopes.
\#Chunks counts all chunks in the retrieval scopes before first-stage
retrieval, and Tok./chunk is the mean chunk length under the Qwen3
tokenizer. All is the mean number of gold chunks per question
(Appendix~\ref{app:labels}); In-pool is the mean number within the
top-50 candidates,
computed over questions with at least one such chunk.}
\label{tab:benchmarks}
\end{table*}

Table~\ref{tab:benchmarks} summarizes the evaluation benchmarks.
HotpotQA, 2WikiMultihopQA, MuSiQue,
NQ, and PopQA used the passages released by HippoRAG~2
\citep{pmlr-v267-gutierrez25a}, and each
dataset shared a single corpus across its questions. The other four
datasets restricted retrieval to the book, novel, conversation, or chat
history associated with each question; in LongMemEval-S, every
question comes with its own chat history of about 50 sessions.
NarrativeQA books were split into non-overlapping 200-token chunks, and
DetectiveQA novels were packed paragraph by paragraph into chunks of at
most 200 tokens. LoCoMo conversations were packed turn by turn into
chunks of at most 300 tokens without crossing session boundaries, and
LongMemEval-S histories were split into two-turn windows with each
message truncated to 200 tokens. Dialogue chunks were prefixed with
their session date. Following QRRanker
\citep{li2026queryfocusedmemoryawarererankerlong}, NarrativeQA and
MuSiQue serve as in-domain benchmarks: both contribute questions to
the training mixture of Section~\ref{sec:training-data}, but their
evaluation questions do not overlap with the training questions. The
other seven benchmarks are not used for training.

\subsection{Relevance Labels}
\label{app:labels}

\begin{table*}[t]
\centering
\small
\begin{tabular*}{\textwidth}{@{\extracolsep{\fill}}llll@{}}
\toprule
\textbf{Dataset} & \textbf{Type} & \textbf{Source} &
\textbf{Relevant if the chunk contains}\\
\midrule
\rowcolor{rowblue}[0pt][\tabcolsep]\multicolumn{4}{@{}l}{\textit{Training}}\\
NarrativeQA     & Silver    & GPT-5.4, MiA-RAG procedure & a passage selected by GPT-5.4\\
MuSiQue         & Annotated & dataset annotation          & a supporting paragraph\\
ConvoMem        & Annotated & dataset annotation          & an evidence message\\
QAConv          & Derived   & dataset answers             & a reference answer string\\
FriendsQA       & Annotated & dataset annotation          & the turn holding the answer\\
\midrule
\rowcolor{rowblue}[0pt][\tabcolsep]\multicolumn{4}{@{}l}{\textit{Evaluation}}\\
HotpotQA        & Annotated & dataset annotation          & a supporting paragraph\\
2WikiMultihopQA & Annotated & dataset annotation          & a supporting paragraph\\
MuSiQue         & Annotated & dataset annotation          & a supporting paragraph\\
NQ              & Annotated & HippoRAG~2 release          & a supporting passage\\
PopQA           & Derived   & HippoRAG~2 release          & the page of the question subject or the answer entity\\
NarrativeQA     & Silver    & GPT-5.4, MiA-RAG procedure & a passage selected by GPT-5.4\\
DetectiveQA     & Annotated & dataset annotation          & an annotated clue paragraph\\
LoCoMo          & Annotated & dataset annotation          & an evidence turn\\
LongMemEval-S   & Annotated & dataset annotation          & a message marked as holding the answer\\
\bottomrule
\end{tabular*}
\caption{Type and source of the gold labels. Annotated labels
come from the evidence annotations of each dataset or of the HippoRAG~2
release \citep{pmlr-v267-gutierrez25a}; derived labels are computed
from annotated answers or entities; silver labels are selected by
GPT-5.4 following MiA-RAG
\citep{li2026mindscapeawareretrievalaugmentedgeneration}. A candidate
chunk is relevant if it contains any of the listed evidence.}
\label{tab:labels}
\end{table*}

Every training and evaluation instance consists of a question, its 50
first-stage candidates ordered by first-stage rank, and a binary label
for each candidate. Labels are assigned at the chunk level, and
Table~\ref{tab:labels} lists the type and source of the labels for
each dataset. Throughout the paper, gold labels refer to these labels
of all three types.

\paragraph{Silver labels.}
NarrativeQA provides no evidence annotations, so its labels for
training and evaluation were constructed as silver
labels. Following MiA-RAG
\citep{li2026mindscapeawareretrievalaugmentedgeneration}, the
question, the question with its answer, and the answer each retrieved
the top 10 chunks of the book with Qwen3-Embedding-8B; the merged list
was ranked by votes and truncated to 10 chunks; and GPT-5.4 selected
the chunks relevant to the question and its reference answer with the
MiA-RAG filtering prompt. In the evaluation set, the 79 questions for
which GPT-5.4 selected no chunk were removed.

\subsection{Hardware and Software}
\label{app:environment}

\paragraph{Hardware.}
Unless stated otherwise, all training, evaluation, data construction,
and baseline runs used a single node with eight GPUs, each providing
141\,GB of HBM3e memory.

\paragraph{Software.}
The node ran with NVIDIA driver
580.65.06 and CUDA 13.0. The main environment used Python 3.12.14,
PyTorch 2.13.0, Transformers 5.16.1 and vLLM 0.28.0.

\section{Ablation of Ranking Objectives}
\label{app:ablation-loss}

Table~\ref{tab:ablation-loss} in Section~\ref{sec:ablation-joint}
compares ranking objectives for the 4B model, leading to three
findings.

(1) \emph{A moderate score scale matters.} Raising the score scale
$\alpha$ from 8 to 12 lowers all four metrics, by 0.97 R@3 and 0.70
NDCG@10 points. A larger scale sharpens the softmax in
$\mathcal{L}_{\mathrm{CE}}$ and further widens margins that are already
correct, consistent with the over-sharpening effect discussed in
Section~\ref{sec:contrastive-objective}.

(2) \emph{The Recall@$k$ surrogate and $\mathcal{L}_{\mathrm{CE}}$ are
complementary.} Replacing $\mathcal{L}_{\mathrm{CE}}$ with the surrogate
yields the highest R@3 (76.81), as expected from an objective that
targets the top-3 boundary directly, but slightly lowers R@1, R@5, and
NDCG@10. Because the surrogate saturates once a relevant candidate
crosses a cutoff, it provides little signal for ordering candidates
within the top ranks. Combining the two objectives retains both
signals: every mixed variant achieves higher R@1 and NDCG@10 than either
objective alone.

(3) \emph{The mixed objective is robust to its weight.} Every metric
varies by at most 0.38 points across $\lambda\in\{1,2,4\}$. With
$\lambda=1$, the mixed objective attains the best R@1 and NDCG@10 but
falls slightly below $\mathcal{L}_{\mathrm{CE}}$ in R@3 and R@5. With
$\lambda=2$, it improves over $\mathcal{L}_{\mathrm{CE}}$ on all four
metrics, achieves the best R@5 and the highest R@3 among the mixed
variants, and stays within 0.11 R@3 points of the surrogate alone.

\section{Straight-Through Updates and Generated Traces}
\label{app:st-discussion}

\paragraph{The trace as a medium.}
\method{} is not required to return the reasoning trace to the user and the trace only
serves as the attention query of the scorer. The quality of a trace is
therefore measured by how much attention it leads the RRHeads to place
on relevant candidates, regardless of its linguistic quality or
readability to humans. The question is thus whether straight-through
training makes the sampled discrete traces more useful for ranking.

\paragraph{What the update does.}
Proposition~\ref{prop:st} states what the straight-through update
computes for one sampled trace $\mathbf u=(u_1,\ldots,u_T)$.

\begin{proposition}
\label{prop:st}
For a trace position $t$, let $f_t(\mathbf e)$ be the ranking loss
obtained when the scorer input at position $t$ is set to
$\mathbf e\in\mathbb{R}^{d}$ and all other inputs are kept, and let
$\boldsymbol\delta_t=\nabla f_t\bigl(E[u_t]\bigr)$ and
$\mathbf g_t=E\boldsymbol\delta_t\in\mathbb{R}^{V}$. Define the
first-order loss of placing token $v$ at position $t$ as
\[
\hat f_t(v)=f_t\bigl(E[u_t]\bigr)+\bigl(E[v]-E[u_t]\bigr)^{\top}\boldsymbol\delta_t .
\]
Then, under Eq.~\ref{eq:straight-through}:
\begin{enumerate}[label=(\roman*), leftmargin=1.6em, itemsep=1pt, topsep=2pt]
\item the forward pass evaluates the loss of the discrete trace,
$\mathcal{L}_{\mathrm{rank}}=f_t\bigl(E[u_t]\bigr)$ for every $t$;
\item with $\boldsymbol\delta_t$ held constant, the straight-through
gradient with respect to the logits $\boldsymbol\ell_t$ satisfies
\[
\begin{aligned}
\frac{\partial\mathcal{L}_{\mathrm{rank}}}{\partial\boldsymbol\ell_t}
&=\nabla_{\boldsymbol\ell_t}\,\mathbb{E}_{v\sim\mathbf p_t}\bigl[\hat f_t(v)\bigr]\\
&=\frac{1}{\tau_{\mathrm{ST}}}\,
\mathbf p_t\odot\bigl(\mathbf g_t-(\mathbf p_t^{\top}\mathbf g_t)\mathbf 1\bigr).
\end{aligned}
\]
\end{enumerate}
\end{proposition}

\begin{proof}
(i) The last two terms of Eq.~\ref{eq:straight-through} are
numerically equal, so
$\widetilde{\mathbf e}_t=E[u_t]+E^{\top}\mathbf p_t-\operatorname{sg}(E^{\top}\mathbf p_t)=E[u_t]$
for every $t$, and the scorer reads exactly the embeddings of the
sampled tokens. Hence
$\boldsymbol\delta_t=\partial\mathcal{L}_{\mathrm{rank}}/\partial\widetilde{\mathbf e}_t$.
(ii) By the chain rule, Eq.~\ref{eq:st-backward}, and the symmetry of
$\mathbf J_t$,
$\partial\mathcal{L}_{\mathrm{rank}}/\partial\boldsymbol\ell_t
=(E^{\top}\mathbf J_t)^{\top}\boldsymbol\delta_t
=\mathbf J_t\mathbf g_t$.
Since $\sum_v p_{t,v}=1$ and $g_{t,v}=E[v]^{\top}\boldsymbol\delta_t$,
\[
\mathbb{E}_{v\sim\mathbf p_t}\bigl[\hat f_t(v)\bigr]
=f_t\bigl(E[u_t]\bigr)-g_{t,u_t}+\mathbf p_t^{\top}\mathbf g_t ,
\]
which depends on $\boldsymbol\ell_t$ only through $\mathbf p_t$. Its
gradient is therefore
$(\partial\mathbf p_t/\partial\boldsymbol\ell_t)^{\top}\mathbf g_t=\mathbf J_t\mathbf g_t$,
and expanding
$\mathbf J_t=\tau_{\mathrm{ST}}^{-1}[\operatorname{Diag}(\mathbf p_t)-\mathbf p_t\mathbf p_t^{\top}]$
gives the last expression.
\end{proof}

Summed over positions, the generator is thus updated by gradient
descent on $\sum_{t}\mathbb{E}_{v\sim\mathbf p_t}[\hat f_t(v)]$, where
$\mathbf p_t=p_{\theta}(\cdot\mid\mathbf x,\mathbf u_{<t})$ is computed
under teacher forcing. Each step raises the logits of tokens whose
$g_{t,v}$ is below the $\mathbf p_t$-weighted average, and the factor
$\mathbf p_t$ concentrates the update on tokens that the generator
already considers likely. Two approximations separate this surrogate
from the objective in Eq.~\ref{eq:joint-objective}: $\hat f_t$ is a
first-order expansion that is accurate only for tokens whose embeddings
lie close to $E[u_t]$, and each $\hat f_t$ holds the other trace tokens
fixed, ignoring how changing $u_t$ would alter the rest of the
continuation and never comparing alternative complete sequences. The
estimator is therefore biased, and decreasing the surrogate does not
guarantee better discrete traces.

\section{Training the Generator with Reinforcement Learning}
\label{app:rl}

Straight-through training passes the gradient of the ranking loss to
the generator through the scorer. A common alternative treats the
scorer as a black-box reward model and trains the generator with
reinforcement learning (RL). This appendix compares straight-through training with RL for
training the generator of \method{} on Qwen3-4B.

\begin{figure}[t]
\centering
\includegraphics[width=\columnwidth]{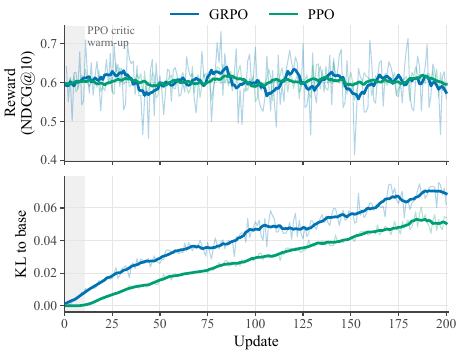}
\caption{Training curves of GRPO and PPO for the Qwen3-4B generator.
The reward is the NDCG@10 that the frozen scorer assigns to each
sampled trace. Thin lines show the value at each update, and thick
lines show an 11-update moving average. The shaded region marks the
first 10 PPO updates, which train only the critic.}
\label{fig:rl-curves}
\end{figure}

\paragraph{Two-stage setup.}
RL requires a fixed reward, so the scorer and the generator were
trained in two stages. In the first stage, the generator was kept at
the base Qwen3-4B weights, and only the scorer was trained on its
reasoning traces, following the training setup of the main
experiments.
In the second stage, this scorer was frozen and served as the reward
model: the reward of a sampled trace $\mathbf{u}$ is the NDCG@10 of the
ranking that the scorer produces from $\mathbf{u}$.

\paragraph{Prompt pool.}
Over all prompts, greedy traces
already reached a mean reward of 0.920, the best of eight samples
improved it by only 0.019, and 73\% of the eight-sample groups received
identical rewards, which leaves group-normalized advantages without a
learning signal. Among prompts with a greedy reward below 0.9, the
greedy reward was 0.591, the best-of-eight gain rose to 0.093, and only
18\% of the groups were tied. RL therefore used the 6,633 prompts of
the training mixture whose greedy reward is below 0.9.

\paragraph{RL algorithms.}
We trained the generator with GRPO \citep{shao2024deepseekmath} and
PPO \citep{schulman2017proximal}, both for 200 updates of 128 episodes
each. Both used a learning rate of $10^{-6}$ with
a clipping ratio of 0.2, two epochs over two minibatches per update,
and a KL coefficient of $\beta=0.04$ toward the base generator.
Rollouts were sampled at temperature 1 in thinking mode with at most
1,024 tokens, the trace budget of the main experiments. GRPO drew eight
samples for each of 16 prompts per update. PPO drew one sample for each of 128
prompts per update and trained a separate critic initialized from
Qwen3-4B, using generalized advantage estimation
\citep{schulman2016gae} with $\lambda=1$ for both the actor and the
critic. Its KL term entered the reward as a per-token penalty, and its
first 10 updates trained only the critic.

\begin{table}[t]
\centering
\footnotesize
\setlength{\tabcolsep}{2pt}
\begin{tabular*}{\columnwidth}{@{\extracolsep{\fill}}lcccc@{}}
\toprule
\textbf{Setting} & \textbf{R@1} & \textbf{R@3} & \textbf{R@5} & \textbf{N@10}\\
\midrule
Scorer only & 43.60\hphantom{\dpos{0.00}} & 76.35\hphantom{\dpos{0.00}} & 83.54\hphantom{\dpos{0.00}} & 81.83\hphantom{\dpos{0.00}}\\
\midrule
\quad + PPO  & 43.66\dpos{0.06} & 76.38\dpos{0.03} & 83.69\dpos{0.15} & 81.90\dpos{0.07}\\
\quad + GRPO & 43.57\dneg{0.03} & 76.31\dneg{0.04} & 83.55\dpos{0.01} & 81.77\dneg{0.06}\\
\midrule
Joint training & \textbf{44.20}\dpos{0.60} & \textbf{76.70}\dpos{0.35} & \textbf{84.01}\dpos{0.47} & \textbf{82.23}\dpos{0.40}\\
\bottomrule
\end{tabular*}
\caption{Macro-9 scores (\%) of \method{} (4B) when the generator is
trained with RL against the frozen scorer. Scorer only is the setting
of Figure~\ref{fig:ablation-joint} in which only the scorer is trained,
and joint training is the setting in which both are trained with the
straight-through estimator, as in \method{}.
Subscripts give the difference from scorer only, with gains in green
and losses in red.}
\label{tab:rl}
\end{table}

\paragraph{Evaluation.}
Table~\ref{tab:rl} compares the RL-trained generators with the
scorer-only setting of Figure~\ref{fig:ablation-joint}, which
corresponds to the first stage before RL. Neither algorithm improves on
it noticeably: PPO changes the four Macro-9 metrics by at most 0.15
points, and GRPO by at most 0.06 points, with slight losses on three of
them. The training curves in Figure~\ref{fig:rl-curves} show the same
picture: the reward of both algorithms stays around 0.60 throughout the
200 updates, while the KL divergence from the base generator grows
steadily to about 0.07 for GRPO and 0.05 for PPO. RL thus moves the
generator away from its initialization without making its traces more
useful to the frozen scorer. In contrast, joint straight-through
training improves over scorer-only training by 0.35 R@3 and 0.40
NDCG@10 points (Section~\ref{sec:ablation-joint}).

\paragraph{Why RL brings little gain.}
Three factors limit RL in this two-stage setting. 
First, analyses of
RL with verifiable rewards find that it mainly concentrates probability
on high-reward outputs that the base model can already sample, rather
than expanding what the model can produce \citep{yue2025does}. A trace that scores higher on one prompt need not reflect a way of
writing that also helps on other prompts. Without a consistent pattern to reinforce, the updates change the
generator without raising the reward: the KL divergence from the base
generator keeps growing while the reward stays flat
(Figure~\ref{fig:rl-curves}).
Second, the frozen scorer was trained on traces of the base generator
and therefore relies on features of their style. It rewards traces that
express these features but not new ways of writing that it has never
learned to read. 
Third, the scorer itself is imperfect: on the same
prompts, even the best of eight traces reaches a mean reward of only
0.684, far below the maximum of 1, so the attainable reward is often
limited by the scorer rather than by the trace. 
Together, these factors
indicate that alternating between training the scorer and training the
generator against it is limited in improving how the two models work
together, whereas straight-through training updates both jointly so
that each adapts to the other.

\begin{table*}[t]
\centering
\small
\setlength{\tabcolsep}{4pt}
\begin{tabular}{llccccc@{\hspace{10pt}}}
\toprule
\textbf{Generator} & \textbf{Method} & \textbf{Single-hop} &
\textbf{Multi-hop} & \textbf{Temporal} & \textbf{Open-domain} &
\textbf{Overall F1}\\
\midrule
GPT-4o-mini & A-Mem~\citep{xu2025amem}$^{\dagger}$ & 44.65 & 27.02 & 45.85 & 12.14 & 39.65\\
 & MemoryOS~\citep{kang-etal-2025-memory}$^{\dagger}$ & 48.62 & 35.27 & 41.15 & 20.02 & 42.84\\
 & Zep~\citep{rasmussen2025zeptemporalknowledgegraph}$^{\dagger}$ & 49.56 & 35.74 & 42.00 & 19.37 & 43.56\\
 & Mem0~\citep{chhikara2025mem0buildingproductionreadyai}$^{\dagger}$ & 47.65 & 38.72 & 48.93 & 28.64 & 45.09\\
 & Nemori~\citep{ma2026deservesmemoryadaptivememory}$^{\dagger}$ & 46.33 & 32.36 & 55.99 & \textbf{29.19} & 44.72\\
 & LightMem~\citep{fang2026lightmemlightweightefficientmemoryaugmented}$^{\dagger}$ & 47.64 & 32.11 & 53.79 & 26.14 & 44.73\\
 & TiMem~\citep{li2026timemtemporalhierarchicalmemoryconsolidation}$^{\dagger}$ & -- & -- & -- & -- & 54.40\\
 & Synapse~\citep{jiang2026synapseempoweringllmagents}$^{\dagger}$ & 48.90 & 35.70 & 50.10 & 25.90 & 40.50\\
 & Membox~\citep{tao2026memboxweavingtopiccontinuity}$^{\dagger}$ & 60.09 & 39.88 & 58.03 & 27.96 & 53.10\\
 & CompassMem~\citep{hu2026memorymattersmoreeventcentric}$^{\dagger}$ & 57.36 & 38.84 & 57.96 & 26.61 & 52.18\\
 & ES-Mem~\citep{zou2026esmemeventsegmentationbasedmemory}$^{\dagger}$ & 50.07 & 36.52 & 47.90 & 24.77 & 45.56\\
 & SFT-Emb-8B~\citep{li2026queryfocusedmemoryawarererankerlong}$^{\dagger}$ & 57.22 & 37.06 & 56.27 & 29.11 & 51.58\\
\cmidrule(l){2-7}
 & Qwen3-Emb-8B (no reranking) & 46.29 & 24.98 & 35.37 & 20.17 & 38.58\\
 & QRRanker (4B) & \textbf{64.78} & 41.65 & 59.06 & 24.92 & 57.02\\
\cmidrule(l){2-7}
 & \method{} (0.6B) & 63.73 & 37.18 & 59.08 & 28.58 & 55.84\\
 & \method{} (4B) & 64.39\rlap{\dneg{0.39}} & \textbf{42.03}\rlap{\dpos{0.38}} & \textbf{59.84}\rlap{\dpos{0.78}} & 26.79\rlap{\dpos{1.87}} & 57.15\rlap{\dpos{0.13}}\\
 & \method{} (8B) & \textbf{64.78} & 41.69 & 59.80 & 28.26 & \textbf{57.37}\\
\midrule
GPT-5-mini & Qwen3-Emb-8B (no reranking) & 46.20 & 27.81 & 38.87 & 29.14 & 40.30\\
 & QRRanker (4B) & 63.55 & 40.16 & 64.16 & 33.18 & 57.62\\
\cmidrule(l){2-7}
 & \method{} (0.6B) & 63.19 & 38.90 & 63.27 & 34.00 & 57.05\\
 & \method{} (4B) & 63.90\rlap{\dpos{0.35}} & \textbf{42.29}\rlap{\dpos{2.13}} & \textbf{66.31}\rlap{\dpos{2.15}} & \textbf{37.15}\rlap{\dpos{3.97}} & \textbf{58.88}\rlap{\dpos{1.26}}\\
 & \method{} (8B) & \textbf{64.13} & 41.77 & 64.76 & 34.34 & 58.43\\
\midrule
GPT-6-luna & Qwen3-Emb-8B (no reranking) & 47.83 & 27.57 & 37.62 & 25.11 & 40.65\\
 & QRRanker (4B) & 64.14 & 43.03 & 66.79 & 31.71 & 58.93\\
\cmidrule(l){2-7}
 & \method{} (0.6B) & 62.98 & 38.85 & 64.05 & \textbf{34.65} & 57.13\\
 & \method{} (4B) & 64.26\rlap{\dpos{0.12}} & 43.19\rlap{\dpos{0.16}} & 66.17\rlap{\dneg{0.62}} & 34.54\rlap{\dpos{2.83}} & 59.06\rlap{\dpos{0.13}}\\
 & \method{} (8B) & \textbf{64.71} & \textbf{43.30} & \textbf{67.26} & 31.41 & \textbf{59.37}\\
\bottomrule
\end{tabular}
\caption{Downstream question answering on LoCoMo (F1, \%) with
different generator LLMs. $^{\dagger}$Results reported by
\citet{li2026queryfocusedmemoryawarererankerlong}, where ``--'' marks
metrics not reported. All non-${\dagger}$ rows rerank the same
Qwen3-Embedding-8B candidates and are evaluated by us. Subscripts on
\method{} (4B) give the difference from QRRanker (4B) under the same
generator. Bold marks the best result in each column within each
generator.}
\label{tab:memory-qa}
\end{table*}

\section{Downstream Question Answering on LoCoMo}
\label{app:memory-qa}

\paragraph{Setup.}
We followed the LoCoMo question answering protocol of
\citet{li2026queryfocusedmemoryawarererankerlong}.
For each question, a reranker ordered the same 50 Qwen3-Embedding-8B
candidates as in Section~\ref{sec:benchmarks}, and only the top-3
chunks were passed to the generator as context, without summaries or
other memory mechanisms. We used GPT-4o-mini, GPT-5-mini, and GPT-6-luna
as generators and report token-level F1 against the reference answers
on four question categories. 
Figure~\ref{fig:locomo-prompt} shows the prompt given to the generator.

\paragraph{Effect of reranking.}
Reranking the same candidates raises overall F1 over the first-stage
order by 16.48--18.79 points across rerankers and generators, whereas
changing the generator moves the overall F1 of any fixed ranking by at
most 2.07 points. In this setting, which chunks reach the generator
matters far more than which generator reads them.

\paragraph{Comparison with memory frameworks.}
With GPT-4o-mini, three \method{} variants achieve higher overall
F1 than every memory framework while passing only three dialogue
chunks to the generator. \method{} (8B) reaches 57.37, 2.97 points
above TiMem, and \method{} (0.6B) still leads
it by 1.44 points. The 4B and 8B variants score higher than all
frameworks on single-hop, multi-hop, and temporal questions.

\paragraph{Comparison with QRRanker.}
At the 4B scale, \method{} achieves higher overall F1 than QRRanker
under all three generators, by 1.26 points with GPT-5-mini and 0.13
points with the other two. The gains concentrate on open-domain
questions (1.87--3.97 points), while single-hop differences stay within
0.39 points. The small overall margins are consistent with the close
LoCoMo retrieval scores of the two rerankers (88.13 vs.\ 88.43 R@3 and
88.96 vs.\ 88.56 NDCG@10; Table~\ref{tab:main-dialogue}).

\begin{figure*}[t]
\begin{promptbox}{LoCoMo Question Answering Prompt}
You are an intelligent memory assistant tasked with retrieving accurate
information from conversation memories.

\medskip
\textbf{CONTEXT:}\\
You have access to memories from two speakers in a conversation. These
memories contain timestamped information that may be relevant to
answering the question.

\medskip
\textbf{INSTRUCTIONS:}
\begin{enumerate}[label=\arabic*., leftmargin=1.5em, nosep]
\item Carefully analyze all provided memories from both speakers.
\item Pay special attention to the timestamps to determine the answer.
\item If the question asks about a specific event or fact, look for
direct evidence in the memories.
\item If the memories contain contradictory information, prioritize the
most recent memory.
\item If there is a question about time references (like ``last year'',
``two months ago'', etc.), calculate the actual date based on the
memory timestamp. For example, if a memory from 4 May 2022 mentions
``went to India last year,'' then the trip occurred in 2021.
\item Always convert relative time references to specific dates,
months, or years. For example, convert ``last year'' to ``2022'' or
``two months ago'' to ``March 2023'' based on the memory timestamp.
Ignore the reference while answering the question.
\item Focus only on the content of the memories from both speakers. Do
not confuse character names mentioned in memories with the actual users
who created those memories.
\item The answer should be less than 5-6 words.
\end{enumerate}

\medskip
\textbf{APPROACH (THINK STEP BY STEP):}
\begin{enumerate}[label=\arabic*., leftmargin=1.5em, nosep]
\item First, examine all memories that contain information related to
the question.
\item Examine the timestamps and content of these memories carefully.
\item Look for explicit mentions of dates, times, locations, or events
that answer the question.
\item If the answer requires calculation (e.g., converting relative
time references), show your work.
\item Formulate a precise, concise answer based solely on the evidence
in the memories.
\item Double-check that your answer directly addresses the question
asked.
\item Ensure your final answer is specific and avoids vague time
references.
\end{enumerate}

\medskip
\textbf{RELEVANT MEMORIES:}\\
\promptvar{context}

\medskip
\textbf{QUESTION:} \promptvar{question}\\
\textbf{ANSWER:}
\end{promptbox}
\caption{Generation prompt for LoCoMo question answering.
\promptvar{context} is filled with the top-3 reranked chunks and
\promptvar{question} with the question.}
\label{fig:locomo-prompt}
\end{figure*}

\section{Effect of Trace Budget}
\label{app:trace-budget}

\begin{figure*}[t]
\centering
\includegraphics[width=\textwidth]{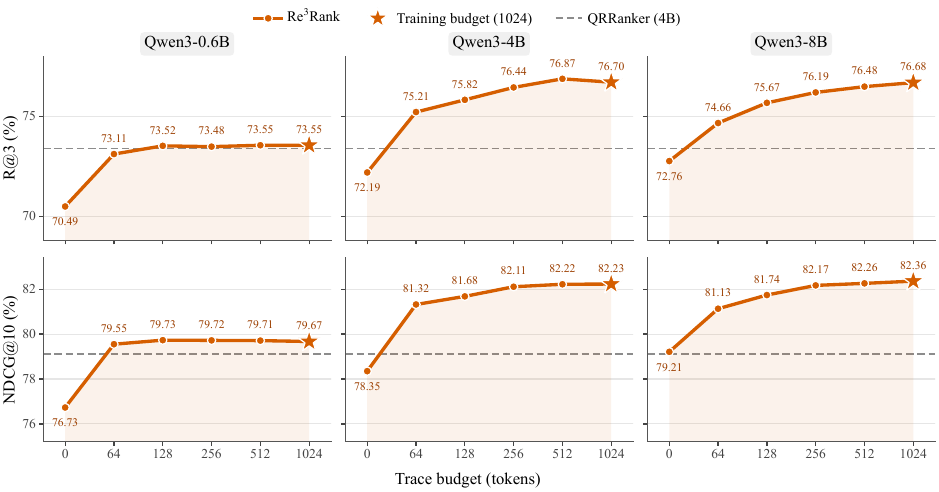}
\caption{Macro-9 R@3 (top) and NDCG@10 (bottom) of \method{} when the
reasoning trace is truncated to a given token budget at inference. All
models were trained with a budget of 1024 tokens. A budget of 0 generates
no trace and uses the question as the attention query. The dashed line
marks QRRanker (4B).}
\label{fig:trace-budget}
\end{figure*}

\begin{figure*}[t]
\centering
\includegraphics[width=\textwidth]{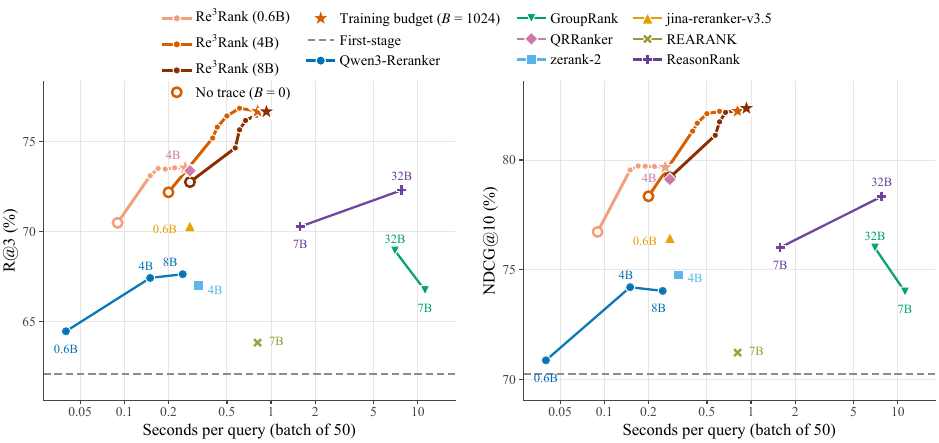}
\caption{Macro-9 R@3 (left) and NDCG@10 (right) against reranking time
per query at a batch size of 50, extending Figure~\ref{fig:metric-vs-latency}
with the trace budgets of Figure~\ref{fig:trace-budget}. Each
\method{} curve connects budgets of 0, 64, 128, 256, 512, and 1024
tokens from left to right. Open circles mark $B=0$, where no trace is
generated and the question serves as the attention query, and stars
mark the training budget of 1024 tokens.
Baseline points repeat Figure~\ref{fig:metric-vs-latency}.}
\label{fig:budget-latency}
\end{figure*}

To examine how much of the reasoning trace \method{} needs, we kept the
trained models fixed and truncated the trace to $B$ tokens at inference
(Figure~\ref{fig:trace-budget}). All models were trained with $B=1024$,
and $B=0$ uses the question as the attention query without generating
a trace.

Compared with the full budget of 1024 tokens, scoring with the question
alone ($B=0$) lowers R@3 by 3.06, 4.51, and 3.92 points for the 0.6B,
4B, and 8B models, respectively. Since the model weights are identical
in both settings, this gap measures how much the trace contributes as
the attention query. Short traces close most of the gap: with 64
tokens, the three models recover 85.6\%, 67.0\%, and 48.5\% of it, and
the 4B and 8B models already surpass QRRanker (4B), which the 0.6B
model does at 128 tokens.

The budget at which accuracy saturates grows with model size: the 0.6B
model levels off at 128 tokens, the 4B model between 256 and 512
tokens, and the 8B model keeps improving up to 1024 tokens. Smaller
budgets therefore reduce generation cost at a small loss in accuracy
(Figure~\ref{fig:budget-latency});
at 256 tokens, the 4B and 8B models trail the full budget by only 0.26
and 0.49 R@3 points.

\section{Results with Gemma~4 Backbones}
\label{app:gemma}

\begin{table*}[t]
\centering
\small
\setlength{\tabcolsep}{5pt}
\begin{tabular}{lcccccccc}
\toprule
& \multicolumn{4}{c}{$\mathcal{L}_{\mathrm{CE}}$} & \multicolumn{4}{c}{$\mathcal{L}_{\mathrm{rank}}$}\\
\cmidrule(lr){2-5}\cmidrule(lr){6-9}
\textbf{Dataset} & \textbf{R@1} & \textbf{R@3} & \textbf{R@5} & \textbf{N@10} & \textbf{R@1} & \textbf{R@3} & \textbf{R@5} & \textbf{N@10}\\
\midrule
\rowcolor{rowblue}\multicolumn{9}{c}{\textit{Gemma-4-E2B}}\\
\midrule
HotpotQA & 49.30 & 96.65 & 98.60 & 97.13 & 49.85 & 97.35 & 98.95 & 97.57\\
2WikiMultihopQA & 54.40 & 98.55 & 99.54 & 99.10 & 54.77 & 99.26 & 99.82 & 99.65\\
MuSiQue & 44.57 & 82.47 & 90.12 & 89.20 & 47.48 & 84.11 & 90.92 & 91.29\\
NQ & 26.35 & 57.77 & 77.52 & 83.31 & 26.48 & 58.68 & 77.44 & 83.58\\
PopQA & 54.80 & 94.65 & 98.90 & 83.15 & 57.65 & 93.55 & 97.65 & 84.21\\
NarrativeQA & 24.30 & 49.36 & 64.00 & 62.99 & 24.69 & 49.79 & 64.50 & 63.59\\
DetectiveQA & 10.40 & 23.90 & 33.11 & 34.08 & 10.08 & 22.37 & 32.94 & 33.70\\
LoCoMo & 72.49 & 88.24 & 92.37 & 88.96 & 72.42 & 88.48 & 92.34 & 89.02\\
LongMemEval-S & 51.65 & 84.74 & 91.89 & 89.69 & 52.55 & 85.38 & 92.33 & 90.59\\
\cmidrule(lr){1-9}
Macro-9 & 43.14 & 75.15 & 82.89 & 80.85 & \textbf{44.00} & \textbf{75.44} & \textbf{82.99} & \textbf{81.47}\\
\midrule
\rowcolor{rowblue}\multicolumn{9}{c}{\textit{Gemma-4-E4B}}\\
\midrule
HotpotQA & 49.45 & 97.80 & 99.20 & 97.52 & 50.00 & 97.90 & 99.25 & 97.82\\
2WikiMultihopQA & 54.80 & 99.42 & 100.00 & 99.76 & 54.70 & 99.57 & 99.95 & 99.72\\
MuSiQue & 46.78 & 86.78 & 93.20 & 92.38 & 46.80 & 86.84 & 93.11 & 92.50\\
NQ & 27.21 & 59.85 & 79.23 & 84.53 & 27.01 & 59.97 & 78.02 & 84.34\\
PopQA & 53.55 & 94.00 & 97.90 & 82.52 & 55.75 & 93.90 & 98.45 & 83.36\\
NarrativeQA & 28.51 & 54.06 & 68.32 & 68.09 & 28.55 & 55.26 & 68.42 & 68.55\\
DetectiveQA & 10.75 & 23.91 & 34.86 & 34.75 & 10.35 & 23.50 & 32.71 & 34.46\\
LoCoMo & 73.75 & 88.80 & 92.61 & 89.73 & 74.03 & 90.01 & 93.25 & 90.29\\
LongMemEval-S & 51.92 & 85.28 & 92.03 & 90.15 & 53.41 & 87.31 & 93.42 & 91.79\\
\cmidrule(lr){1-9}
Macro-9 & 44.08 & 76.66 & \textbf{84.15} & 82.16 & \textbf{44.51} & \textbf{77.14} & 84.06 & \textbf{82.54}\\
\bottomrule
\end{tabular}
\caption{Per-dataset reranking results of \method{} with Gemma~4
backbones (\%). Metrics follow Table~\ref{tab:main-wiki}. In the Macro-9
rows, bold marks the better objective on each metric.}
\label{tab:gemma-full}
\end{table*}

We applied the training method of Section~\ref{sec:method} to Gemma-4-E2B
and Gemma-4-E4B \citep{gemmateam2026gemma4}, which have 2.3B and 4.5B effective
parameters (5.1B and 8B including embeddings). RRHeads were selected as
in Section~\ref{sec:rrhead-selection}, giving $\{(34,2),(9,7)\}$ for
Gemma-4-E2B and $\{(35,0),(29,4)\}$ for Gemma-4-E4B. As Table~\ref{tab:gemma} shows,
Gemma-4-E2B falls between Qwen3-0.6B and Qwen3-4B on all four Macro-9
metrics, and Gemma-4-E4B slightly exceeds Qwen3-4B and Qwen3-8B in R@3
and NDCG@10. Both Gemma~4 variants also outperform every baseline in
Section~\ref{sec:main-results} on all four Macro-9 metrics.

Table~\ref{tab:gemma-full} reports per-dataset results, together with
the same backbones trained with $\mathcal{L}_{\mathrm{CE}}$ alone. The
mixed objective improves Macro-9 R@3 and NDCG@10 over
$\mathcal{L}_{\mathrm{CE}}$ on both backbones and achieves higher R@3
on seven of the nine datasets for each.

\section{What the Reasoning Trace Contributes}
\label{app:trace-content}

\begin{table*}[t]
\centering
\small
\begin{tabular*}{\textwidth}{@{\extracolsep{\fill}}lrcccc c@{}}
\toprule
& & \multicolumn{2}{c}{\textbf{As is}} &
\multicolumn{2}{c}{\textbf{At trace length}} & \\
\cmidrule(lr){3-4}\cmidrule(lr){5-6}
\textbf{Query text after \texttt{<think>}} & \textbf{Tokens} &
\textbf{R@3} & \textbf{N@10} & \textbf{R@3} & \textbf{N@10} &
\textbf{$\Delta$R@3 [95\% CI]}\\
\midrule
\rowcolor{rowblue}[0pt][\tabcolsep]\multicolumn{7}{@{}l}{\textit{Question and neutral text ($\Delta$R@3 vs.\ full trace)}}\\
Question in the user turn ($B=0$) & 15  & 72.19 & 78.35 & --    & --    & $-4.51$ [$-5.00$, $-4.03$]\\
Question                          & 15  & 75.27 & 81.24 & 75.57 & 81.07 & $-1.13$ [$-1.46$, $-0.80$]\\
Neutral sentence                  & --  & --    & --    & 74.50 & 80.22 & $-2.20$ [$-2.57$, $-1.84$]\\
\midrule
\rowcolor{rowblue}[0pt][\tabcolsep]\multicolumn{7}{@{}l}{\textit{Content extracted from the trace ($\Delta$R@3 vs.\ full trace)}}\\
Answer only                       & 6   & 71.60 & 78.18 & 74.64 & 80.32 & $-2.05$ [$-2.43$, $-1.69$]\\
Chunk identifiers                 & 68  & 74.51 & 80.06 & 75.15 & 80.62 & $-1.55$ [$-1.89$, $-1.23$]\\
Keywords                          & 111 & 74.39 & 80.03 & 75.17 & 80.67 & $-1.53$ [$-1.87$, $-1.20$]\\
Verbatim excerpts                 & 249 & 75.40 & 80.88 & 75.38 & 80.91 & $-1.32$ [$-1.64$, $-0.99$]\\
Evidence summary                  & 47  & 75.63 & 81.26 & 75.89 & 81.39 & $-0.81$ [$-1.12$, $-0.52$]\\
Question + all trace entities     & 76  & 75.30 & 80.83 & --    & --    & $-1.40$ [$-1.78$, $-1.01$]\\
\midrule
\rowcolor{rowblue}[0pt][\tabcolsep]\multicolumn{7}{@{}l}{\textit{Perturbed traces ($\Delta$R@3 vs.\ full trace)}}\\
Trace of another question         & 881 & 70.82 & 77.04 & --    & --    & $-5.88$ [$-6.36$, $-5.40$]\\
Trace without verbatim excerpts   & 535 & 76.66 & 82.20 & --    & --    & $-0.04$ [$-0.23$, $+0.15$]\\
Trace without chunk identifiers   & 810 & 76.65 & 82.15 & --    & --    & $-0.05$ [$-0.18$, $+0.09$]\\
\midrule
\rowcolor{rowblue}[0pt][\tabcolsep]\multicolumn{7}{@{}l}{\textit{Required entities ($\Delta$R@3 vs.\ full trace)}}\\
Trace without required entities   & 844 & 76.11 & 81.68 & --    & --    & $-0.59$ [$-0.73$, $-0.45$]\\
Question + named required entities & 20 & 77.78 & 83.33 & --    & --    & $+1.08$ [$+0.72$, $+1.43$]\\
Question + as many other trace entities & 19 & 74.98 & 80.85 & -- & -- & $-1.71$ [$-2.10$, $-1.33$]\\
Question + all required entities  & 22  & 80.71 & 86.05 & --    & --    & $+4.01$ [$+3.57$, $+4.47$]\\
\midrule
\textbf{Full trace}               & 881 & \textbf{76.70} & \textbf{82.23} & -- & -- & --\\
\midrule
\rowcolor{rowblue}[0pt][\tabcolsep]\multicolumn{7}{@{}l}{\textit{Untrained Qwen3-4B scorer with the same RRHeads ($\Delta$R@3 from training)}}\\
Question in the user turn ($B=0$) & 15  & 70.01 & 76.60 & --    & --    & $+2.18$ [$+1.65$, $+2.73$]\\
Question                          & 15  & 71.97 & 77.96 & 64.03 & 73.52 & $+11.54$ [$+10.94$, $+12.14$]\\
Neutral sentence                  & --  & --    & --    & 60.37 & 70.87 & $+14.13$ [$+13.50$, $+14.79$]\\
Full trace                        & 881 & 67.95 & 74.16 & --    & --    & $+8.75$ [$+8.17$, $+9.36$]\\
\bottomrule
\end{tabular*}
\caption{Macro-9 scores (\%) of \method{} (4B) when the query text
read by the scorer is replaced, with the scorer and its RRHeads fixed.
Tokens is the mean length of the text as is; at trace length, the text
is repeated line by line to the token length of the original trace. The
neutral sentence (``Let me look through the retrieved chunks.'') carries
no question-specific content and was only used at trace length. The first row
keeps the question in the user turn, as for $B=0$ in
Appendix~\ref{app:trace-budget}. Required entities are the bridge
entities and answers that the question needs but omits, annotated by
GPT-6-luna from the question, its answer, and its gold chunks; named
ones are those mentioned in the trace. Variants built with them rely
on gold labels and serve only for analysis; the control adds as many
other entities from the trace.
$\Delta$R@3 compares the full trace
with the version at trace length when it exists and with the text as
is otherwise. The last group replaces the trained scorer with the
untrained Qwen3-4B; its last column is the R@3 of the trained scorer
minus that of the untrained one for the same query text, at trace
length when it exists. Intervals come from a paired bootstrap over
questions.}
\label{tab:trace-content}
\end{table*}

We analyze which parts of the reasoning trace drive the ranking of
\method{} (4B). The generator's greedy traces and the trained scorer
were kept fixed, and only the query text after \texttt{<think>} in the
scorer input was replaced. Each variant was used as is and, to compare
variants with the same number of query positions, also repeated line
by line to the token length of the original trace (881 tokens on
average). As a control, the neutral sentence ``Let me look through the
retrieved chunks.'' was repeated in the same way, which adds query
positions without question-specific content. The answer, keywords, and
evidence summary were
extracted from the trace by GPT-6-luna given only the question and the
trace, and verbatim excerpts and chunk identifiers by string matching.
To isolate the effect of training, the question, the neutral sentence,
and the full trace were also scored by the untrained Qwen3-4B with the
same RRHeads, input, and scoring, where the full trace still came from
the trained generator. Table~\ref{tab:trace-content} reports
Macro-9 scores; differences were tested with a paired bootstrap over
questions.

\paragraph{Format, positions, and content.}
Relative to reading the question in the user turn, the full trace
improves R@3 by 4.51 points. Placing the question after
\texttt{<think>}, where the scorer reads its queries during training,
accounts for 3.08 of these points, and the reasoning content adds 1.43
points over this question-only query and 1.13 points over the question
repeated to trace length. Additional query positions alone do not help:
repeating the question changes R@3 by $+0.30$, which is not
significant, and a repeated neutral sentence lowers it by 0.77 points.
The gain also depends on question-specific content, as the trace of
another question falls 5.88 points below the full trace.

\paragraph{Which content matters.}
Each component extracted from the trace, repeated to trace length,
scores below the full trace; the evidence summary comes closest
($-0.81$ R@3). Removing verbatim
excerpts or chunk identifiers from the trace changes R@3 by less than
0.1 points. The ranking signal thus comes from the question-specific
reasoning context as a whole; keywords, verbatim excerpts, and chunk
identifiers alone recover only part of it. Even the neutral sentence,
which carries no content, reaches 74.50 R@3 with the trained scorer,
above the question in the user turn (72.19). We attribute this to the
causal attention of the scorer: \textit{the hidden state at each query position
is built from all preceding tokens, including the question and the
candidates, so its attention query depends mainly on this context and
only partly on the embedding of the token at that position.}

\paragraph{Required entities.}
Naming the required entities is one source of the gain of the trace.
Whether an entity is required was judged by GPT-6-luna from the
question, its answer, and its gold chunks, so the variants built on
this judgment give near-oracle results that are unavailable at
inference. Relative to the full trace, deleting them lowers R@3 by 0.59 and
NDCG@10 by 0.55 points (76.11 vs.\ 76.70 and 81.68 vs.\ 82.23). Since
the full trace outperforms the question after \texttt{<think>} by 1.43
R@3 and 0.99 NDCG@10 points, the deletion accounts for 41\% and 56\% of
this gain, and the trace without required entities still outperforms
the question after \texttt{<think>} (76.11 vs.\ 75.27 R@3). Which
entities are named also matters. Appending the required entities
named in the trace to the question reaches 77.78 R@3 with 20 tokens,
1.08 points above the full trace, whereas appending as many other
entities from the trace (74.98) or all of its entities (75.30) brings
no gain over the question after \texttt{<think>} (75.27).
Appending all required entities, including those the trace misses,
gives an upper bound of 80.71 R@3. The trace thus supplies the
required entities within reasoning that singles them out among the
many entities in the candidates, which a plain entity list achieves
only with this near-oracle selection.

\paragraph{Effect of training.}
The untrained scorer ranks better with the question after
\texttt{<think>} than with the full trace (71.97 vs.\ 67.95 R@3), and
long query text degrades its ranking: repeating the question or the neutral
sentence to trace length lowers R@3 by 7.95 and 11.61 points. Training
mitigates this degradation, as both repeated texts stay within 0.8
points of the question after \texttt{<think>} for the trained scorer.
It also raises the R@3 of the full trace by 8.75 points, so the full
trace moves from below the question after \texttt{<think>} to above it.
The effect of training extends beyond the reasoning trace to filler
text at the query positions: training raises the R@3 of the repeated
question and the neutral sentence by 11.54 and 14.13 points, more than
that of the full trace. This again follows from how the hidden states
are built. \textit{Each query position aggregates the question and the
candidates from its preceding context, and training teaches the scorer
to route attention from these contextual states to relevant
candidates, whichever tokens occupy the positions.}

\section{Comparison with Closed-Source LLM Rerankers}
\label{app:llm-rerankers}

\begin{table}[t]
\centering
\footnotesize
\setlength{\tabcolsep}{2pt}
\begin{tabular*}{\columnwidth}{@{\extracolsep{\fill}}lcccc@{}}
\toprule
\textbf{Reranker} & \textbf{R@1} & \textbf{R@3} & \textbf{R@5} & \textbf{N@10}\\
\midrule
\rowcolor{rowblue}[0pt][\tabcolsep]\multicolumn{5}{@{}l}{\textit{Closed-source LLMs}}\\
GPT-6.1-sol & \textbf{46.85} & \textbf{79.19} & \textbf{86.11} & \textbf{84.87}\\
GPT-6-luna  & 46.15 & 77.88 & 84.67 & 83.55\\
\midrule
\rowcolor{rowblue}[0pt][\tabcolsep]\multicolumn{5}{@{}l}{\method{} \textit{with Qwen3}}\\
0.6B & 42.34 & 73.55 & 81.43 & 79.67\\
4B   & 44.20 & 76.70 & 84.01 & 82.23\\
8B   & 44.12 & 76.68 & \textbf{84.36} & 82.36\\
\midrule
\rowcolor{rowblue}[0pt][\tabcolsep]\multicolumn{5}{@{}l}{\method{} \textit{with Gemma~4}}\\
Gemma-4-E2B & 44.00 & 75.44 & 82.99 & 81.47\\
Gemma-4-E4B & \textbf{44.51} & \textbf{77.14} & 84.06 & \textbf{82.54}\\
\bottomrule
\end{tabular*}
\caption{Macro-9 scores (\%) of closed-source LLMs used as listwise
rerankers and of \method{} with Qwen3 and Gemma~4 backbones. Bold
marks the best result among the closed-source LLMs and among the
versions of \method{}.}
\label{tab:llm-rerankers}
\end{table}

\begin{figure*}[t]
\begin{promptbox}{Listwise Reranking Prompt for Closed-Source LLMs}
\textbf{System:} You are an expert at judging which retrieved passages
are needed to answer a query.

\medskip
\textbf{User:} I will give you 50 passages, each labelled with a number
in square brackets. Rank them by how useful they are for answering this
query: \promptvar{question}

\medskip
\promptvar{passages}

\medskip
Query: \promptvar{question}

\medskip
A passage is useful when it states information that the answer depends
on: the answer itself, or, when reaching the answer takes more than one
step, a fact needed for one of the steps (for example, the passage that
identifies a person, place or work that the query only describes, or
one of several facts that the query compares or combines). Rank first
the passages that state such information, the most essential first. A
passage that only shares the topic or the wording of the query is not
useful.

\medskip
Reply with the numbers of the 10 most useful passages, the most useful
first, in the form \texttt{[4] > [17] > [2] > ...} : exactly 10
different numbers and nothing else.
\end{promptbox}
\caption{Listwise reranking prompt for GPT-6.1-sol and GPT-6-luna.
\promptvar{passages} lists the 50 candidates as \texttt{[1]} to
\texttt{[50]} in first-stage order, separated by blank lines, and
\promptvar{question} is the question.}
\label{fig:llm-rerank-prompt}
\end{figure*}

We compared \method{} with two closed-source reasoning LLMs, GPT-6.1-sol
and GPT-6-luna, used as listwise rerankers. They serve only as a
reference for the ranking quality of frontier models, separate from
the baselines of Section~\ref{sec:baselines}. Each model was called once
per question with the question and all 50 candidates in first-stage
order. Both models received the same prompt
(Figure~\ref{fig:llm-rerank-prompt}), ran at their default reasoning
effort, and returned the identifiers of
their 10 most useful candidates, which were ranked first in that
order, followed by the remaining candidates in first-stage order.
Table~\ref{tab:llm-rerankers} reports Macro-9 scores from a single run.

The gap between \method{} and these frontier LLMs is small.
GPT-6.1-sol and GPT-6-luna lead \method{} (4B) by only 2.49 and 1.19
R@3 points, \method{} with Gemma-4-E4B comes within 0.74 R@3 points of
GPT-6-luna, and \method{} (8B) is on par with GPT-6-luna on R@5 (84.36
vs.\ 84.67; paired difference $+0.31$ [$-0.19$, $+0.81$]). \method{}
reaches this level with open backbones of a few billion parameters that
run locally, at 0.81\,s per query for the 4B model under batched
inference (Figure~\ref{fig:metric-vs-latency}), whereas the two LLMs
are closed-source models of undisclosed size whose API calls took 5.2
and 11.0\,s per question on average. Moreover, when the scorer of
\method{} (4B) reads the question followed by all required entities
(Table~\ref{tab:trace-content}), it reaches 80.71 R@3 and 86.05
NDCG@10, above GPT-6-luna used as a reranker (77.88 and 83.55), the
same model that identified these entities. This result shows that the
attention-based scorer is effective and that \method{} has
considerable potential as its generator learns to state the required
information more precisely.

\end{document}